\documentclass[10pt,twocolumn,letterpaper]{article}
\usepackage[pagenumbers]{cvpr}
\usepackage{graphicx,amsmath,amssymb,booktabs,multirow,xspace,array,longtable,needspace,colortbl}
\definecolor{cvprblue}{rgb}{0.21,0.49,0.74}
\usepackage[breaklinks,colorlinks,allcolors=cvprblue]{hyperref}
\usepackage[capitalize]{cleveref}
\hypersetup{pdftitle={RELAY: Learning Compatible Representations One Encoder at a Time},
            pdfauthor={Yu Wang, Chuan Qin, Qiang Zhang}}
\newcommand{\method}{RELAY\xspace}
\newcommand{\sg}{\operatorname{sg}}
\newcommand{\simc}{\operatorname{cos}}
\title{RELAY: Learning Compatible Representations\\One Encoder at a Time}
\author{%
  \begin{tabular}{@{}ccc@{}}
    Yu Wang & Chuan Qin & Qiang Zhang \\
    \small\texttt{tuozhi.wy@alibaba-inc.com} &
    \small\texttt{chuan.qinqc@alibaba-inc.com} &
    \small\texttt{larmmi.zhang@alibaba-inc.com} \\[3pt]
    \multicolumn{3}{c}{Qwen Business Unit of Alibaba}
  \end{tabular}%
}
\begin{document}
\maketitle
\begin{abstract}
Old and new document embeddings coexist during incremental gallery
re-encoding. \method trains a new query encoder with a learned projection
against the original gallery, then freezes both to train the new document
encoder. The same query vector
scores both document versions throughout a single re-encoding pass.
After backfill, optional query refinement drops the projection and reuses
the stored document vectors.
Across three text-to-image retrieval upgrades on COCO and Flickr30K,
RELAY improves Recall@1 on untouched galleries by 8.66--10.23 percentage
points over the original systems. For direct re-encoding, RELAY ranks first
in both Qwen-to-Qwen upgrades and second in CLIP-to-Qwen by area under the
Recall@1 curve, under both native scoring and matched calibration.
With embeddings held fixed, calibrating scores separately for each
document version can reduce drops in Recall@1 during backfill and
change which method achieves the highest backfill AUC.
After query refinement, final Recall@1 is within 0.97 percentage points
of conventional joint training in all three upgrades.
\end{abstract}

\section{Introduction}
A retrieval system can replace its query encoder before it finishes
re-encoding a large document gallery. This gives the new query encoder
a chance to improve retrieval on the untouched gallery. It must then
rank a mix of old and new document vectors during backfill, while the
fully updated system should recover the quality of a jointly trained pair.

Strong results on the original and fully updated galleries can conceal
a drop during backfill. In our upgrade from CLIP to Qwen3-VL-Embedding 2B,
the retrieval adaptation of BT$^2$~\cite{zhou2023bt2} with native scores
starts at 61.27\% Recall@1 and finishes at 72.32\%. After re-encoding
just 10\% of the gallery, Recall@1 drops to 13.81\%. Calibrating scores
on the same features recovers 61.42\% at that point (\cref{fig:curves}A).
When old and new documents compete for the same ranks, a version that
tends to produce larger scores can displace relevant documents from
the other, even when each version ranks its own documents well.

Backward-compatible training lets an upgraded model search the stored
gallery~\cite{shen2020bct,zhang2022ract,pan2023advbct}. Forward maps convert
stored vectors to the new model's space~\cite{ramanujan2022fct}. Separate
search branches can also merge results from old and new
subgalleries~\cite{seo2025metric}. We ask whether the query that first
serves the old gallery can also guide the training of the new document
encoder, so the same query can serve throughout backfill.

\begin{figure*}[t]\centering
\includegraphics[width=\textwidth]{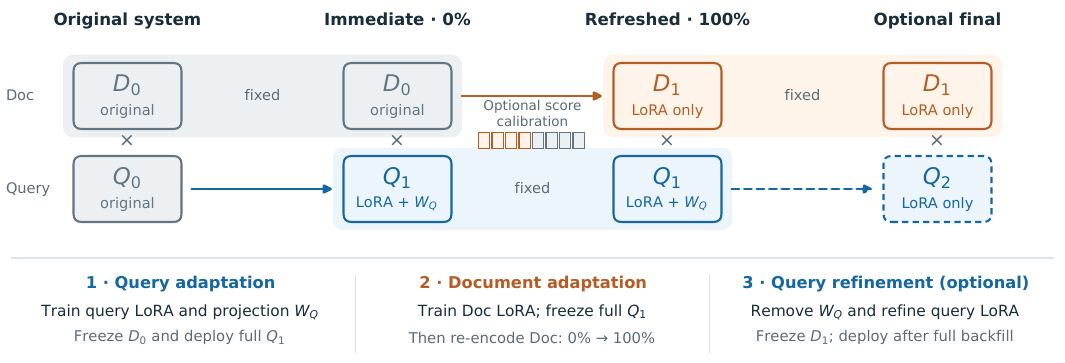}
\caption{\textbf{RELAY uses the deployed query encoder to train the new document encoder.}
The temporary-projection variant is shown. Query adaptation learns $Q_1$
(LoRA and query projection $W_Q$) against the old $D_0$ gallery. Document
adaptation freezes $Q_1$, including $W_Q$, and trains the new document
encoder. For retrieval, $D_1$ takes the first $d$ coordinates of the
encoder output and L2-normalizes the resulting vector, where $d$ is the
dimension of the original gallery embeddings.
After backfill, optional refinement removes the query projection and learns $Q_2$
against the frozen $D_1$. Subscripts count updates within each role.
Continuous bands trace the fixed reference; dashed elements denote optional
refinement. Optional score calibration acts on the mixed-gallery scores.}
\label{fig:lifecycle}\end{figure*}

\method trains in this order (\cref{fig:lifecycle}). We first adapt a query
encoder to the old gallery. We then freeze the entire query function and
train the new document encoder against it. During backfill, the same
query encoder scores both document versions as new vectors replace old
ones in a single pass. A learned query projection helps align
with the old embedding space. The new document encoder uses a fixed prefix
of its embeddings, and optional query refinement removes the learned query
projection once all documents have been re-encoded. The new document
vectors stay in place during refinement. Each stage uses the standard
retrieval loss.

We evaluate Qwen 2B-to-2B, CLIP-to-Qwen 2B, and Qwen 2B-to-8B upgrades
on COCO and Flickr30K. RELAY raises Recall@1 before the first document
is re-encoded in every setting. Serving both versions with one fixed
query encoder, RELAY leads the two Qwen-to-Qwen upgrades in direct re-encoding
AUC and places second in CLIP-to-Qwen under both scoring protocols.
After refinement, the final pair stays within 0.97 percentage points
of Joint, which trains both encoders simultaneously with the same
retrieval loss and without a compatibility constraint.

We make three contributions.
\begin{enumerate}
\item A training procedure that adapts the query encoder to the original
gallery, then freezes it to train the new document encoder. The same query
encoder serves both document versions throughout backfill.
\item A temporary query projection and a controlled comparison of
document projections under the same frozen query encoder.
\item A fixed-feature evaluation of per-version score calibration across
13 methods and both deployment paths. Calibration raises re-encoding AUC
in 33 of 39 method--upgrade pairs.
\end{enumerate}

\section{Related Work}
\paragraph{Compatible encoder training.}
BCT trains new embeddings with the old classifier to preserve retrieval
against an existing gallery~\cite{shen2020bct}. BT$^2$ uses learned basis
transformations and additional dimensions to preserve new-model
information~\cite{zhou2023bt2}. RACT adds a same-instance compatibility
term with old and new negatives to reduce regressions during
backfill~\cite{zhang2022ract}. AdvBCT aligns old and new image embedding
distributions through adversarial training and constrains new embeddings
with elastic boundaries around old class centers~\cite{pan2023advbct}.
These methods build on classification supervision. XBT extends compatible
training to vision--language retrieval by pretraining a projection on text,
then using parameter-efficient image--text training~\cite{jang2025xbt}.

\paragraph{Feature transformations.}
FCT stores side information for future feature
conversion~\cite{ramanujan2022fct}. Embedding-Converter maps stored vectors
to the target space while preserving pairwise
geometry~\cite{yoon2025converter}. FastFill combines forward alignment with
uncertainty-based refresh ordering~\cite{jaeckle2023fastfill}, and
$\lambda$-Orth learns backward and forward maps with geometric
regularization~\cite{ricci2025lambda}. BiCT combines a backward-compatible
encoder for immediate access with a forward map that progressively converts
stored vectors~\cite{su2022bict}. SLERP interpolates the old and aligned new
query~\cite{ricci2026slerp}, while MKP maps new queries to the old space with
reconstruction and relational constraints~\cite{byun2026mkp}.

\paragraph{Ranking mixed galleries.}
MCT uses a metric-compatible contrastive loss to learn maps on top of
a jointly trained target model. Its Rank Merge framework uses different
mapped query vectors for the two gallery versions and merges their
results~\cite{seo2025metric}. RELAY first adapts
the query encoder to the original gallery, then trains the new document
encoder against that frozen query function. It scores both versions with
the same query vector during backfill. We use isotonic
regression~\cite{zadrozny2002calibration} to study score mixing with
embeddings and within-version rankings held fixed.

\paragraph{Updating one retrieval role at a time.}
Atlas adapts a query encoder against a fixed document
index~\cite{izacard2023atlas}. CoGR alternates query and item keyword
generators using reinforcement learning~\cite{dai2026cogr}. Continual
dense retrieval also studies compatibility under changing query
distributions~\cite{goswami2025drift}. RELAY schedules one-sided updates
around a single gallery upgrade. \Cref{tab:method-requirements} compares
training access, gallery updates, mixed-gallery scoring, and endpoint
requirements across methods.

\section{Method}
\subsection{Retrieval during a model upgrade}
Let $Q_0,D_0$ be the old query and document encoders. The gallery
$\mathcal G$ contains $N$ fixed documents, indexed by $j\in\{1,\ldots,N\}$,
with raw content $x_j$ (an image in our experiments). RELAY produces an
updated query encoder $Q_1$ and a document encoder $D_1$. During backfill,
$Q_1$ stays fixed while $D_1(x_j)$ gradually replaces $D_0(x_j)$.

The progress variable $p\in[0,1]$ specifies the fraction to replace.
For a fixed replacement order, $\pi(j)\in\{1,\ldots,N\}$ is document
$j$'s rank in that order, with each rank used once. We replace
$m_p=\lfloor pN\rfloor$ documents, where $\lfloor\cdot\rfloor$ rounds
down to an integer. The active vector of document $j$ is
\begin{equation}
 \widetilde d_j(p)=
 \begin{cases}D_1(x_j),&\pi(j)\le m_p,\\D_0(x_j),&\pi(j)>m_p.\end{cases}
 \label{eq:mixed}
\end{equation}
The first $m_p$ documents use new vectors; the rest retain old vectors.
Increasing $p$ preserves earlier replacements, and each document contributes
one active vector. At $p=0$, retrieval uses $Q_1\times D_0$ on the untouched
gallery. At $p=1$, it uses $Q_1\times D_1$. A subsequent query refinement
produces the separate final pair $Q_2\times D_1$.

Let $P(p)$ be retrieval performance at progress $p$. We report entry
$P(0)$ and the sampled minimum. We compute the area under the backfill
curve (AUC) by trapezoidal integration over the measured fractions. We evaluate the
final refined pair separately. Direct re-encoding starts from the
original vectors. For methods that first convert the stored vectors, we
report a separate AUC for conversion and another for re-encoding the
fully converted gallery. These AUCs measure retrieval against the
fraction completed in each phase. We report deployment costs separately.

\subsection{Query-first training}
The training sequence in \cref{fig:lifecycle} follows from the available
gallery. Before backfill, the new query must retrieve old document vectors.
During backfill, the new documents must work with that deployed query.
We optimize these two pairs in turn.

RELAY has three training stages: (1) query adaptation, (2) document
adaptation, and (3) optional query refinement. They produce $Q_1$, $D_1$,
and $Q_2$, respectively.
Encoder subscripts count updates within each role, with 0 denoting the
original system: $Q_0\to Q_1\to Q_2$ and $D_0\to D_1$.

\paragraph{Retrieval objective.}
Let a batch contain $B_q$ queries $q_i$ and $B_d$ distinct documents $x_j$.
Write $x_i^+$ for query $q_i$'s positive document from this batch.
We use a query-to-document contrastive objective, averaged over a set
$\mathcal K$ of embedding dimensions:
\begin{equation}
 \begin{aligned}
 \mathcal L_{ret}(Q,D)={}&-\frac{1}{B_q|\mathcal K|}
 \sum_{k\in\mathcal K}\sum_{i=1}^{B_q}\\
 &\log\frac{\exp(s_k(Q(q_i),D(x_i^+))/\tau)}
 {\sum_{j=1}^{B_d}\exp(s_k(Q(q_i),D(x_j))/\tau)}.
 \end{aligned}
 \label{eq:retrieval}
\end{equation}
Here $s_k$ is cosine similarity after normalizing the first $k$ dimensions of each embedding,
following Matryoshka training~\cite{kusupati2022matryoshka}, and $\tau$ is
the temperature. Our image--text sampler uses five captions per image,
so $B_q=5B_d$. Each caption has its image as the positive and the other
batch images as negatives. Caption count is a sampling choice, not a
requirement of RELAY.

\paragraph{Training the serving pairs.}
Each stage trains one role against the other role's serving function.
Let $E^Q_t$ and $E^D_t$ denote the encoders before output processing.
The corresponding serving functions $Q_t$ and $D_t$ include that
processing. We write $\operatorname{normalize}(z)=z/\|z\|_2$
and $\operatorname{prefix}_d(z)=z_{1:d}$ for L2 normalization and retaining
the first $d$ dimensions of an embedding, where $d$ is the shared retrieval dimension.
The temporary-projection variant uses a learned query projection $W_Q$
to enter the old document space:
\begin{align}
 Q_1(q)&=\operatorname{normalize}(W_Q E^Q_1(q)),\\
 D_1(x)&=\operatorname{normalize}(\operatorname{prefix}_d(E^D_1(x))),\\
 Q_2(q)&=\operatorname{normalize}(\operatorname{prefix}_d(E^Q_2(q))).
 \label{eq:temporary-functions}
\end{align}
Thus $Q_1$ uses a learned projection, while $D_1$ and $Q_2$ use a fixed
dimension prefix. All three normalize the resulting vector.
The corresponding training objectives are
\begin{align}
 \min_{E^Q_1,W_Q}\quad&\mathcal L_{ret}(Q_1,\sg(D_0)),\\
 \min_{E^D_1}\quad&\mathcal L_{ret}(\sg(Q_1),D_1),\\
 \min_{E^Q_2}\quad&\mathcal L_{ret}(Q_2,\sg(D_1)).
 \label{eq:stages}
\end{align}
We update only the LoRA parameters of each encoder $E$ and freeze its
backbone. The stop-gradient operator $\sg$ holds the opposite serving
function fixed. During document adaptation, we freeze the deployed
$Q_1$, including $W_Q$, and train $D_1$ against it. Once all documents
have been re-encoded with $D_1$, we can remove $W_Q$ and refine the
query LoRA against the stored $D_1$ vectors.

The encoders use independent role LoRAs over a frozen backbone~\cite{hu2022lora}.
Query adaptation starts from the target Query initialization. Document
adaptation uses its own untrained Doc role, not a copy of the trained
Query. Query refinement starts from the adapted Query LoRA.
In the same-base case the target starts from the old Qwen backbone; in
cross-family/capacity upgrades it starts from the corresponding new Qwen
backbone. $W_Q$ has no bias or activation and starts as $[I_d\;0]$. The
query projection reads the full target embedding before the retrieval-prefix
loss. The final encoders use fixed prefixes to match dimensions and retain
their trained LoRAs.

We denote stage choices by $\mathrm{P}_{b_1b_2b_3}$, where $b_i$ indicates
a learned output projection on the role \emph{trained} in stage $i$.
P000 uses only LoRA; P111 retains projections on each trained side;
P100 learns its projection during query adaptation and retains it through
backfill. P110 removes the query projection during refinement.
The frozen side always retains its full inherited function. Optimizers
and schedules restart at stage boundaries. Fixed features can be cached
or evaluated without gradients. Query and document adaptation suffice for
migration; optional query refinement falls outside the backfill AUC.

\subsection{Calibrating mixed-gallery scores}
\label{sec:score-calibration}
A query may rank each document version well and still compare their scores
poorly. For example, systematically larger scores from newly encoded
documents can suppress relevant old documents early in backfill. We use
isotonic regression to fit one monotone score map per document version.

For version $v\in\{0,1\}$, let
$s_v(q,x)=\simc(Q_1(q),D_v(x))$. From the development set (DEV), we
sample weighted score--relevance triples $(s,y,w)$ for version $v$ and
collect them in $\mathcal V_v$. Here $w$ is the sampling weight,
$y=1$ for a relevant document and $y=0$ otherwise. We fit
\begin{equation}
 g_v=\arg\min_{g\;\mathrm{nondecreasing}}
 \sum_{(s,y,w)\in\mathcal V_v}w(g(s)-y)^2.
 \label{eq:isotonic}
\end{equation}
At retrieval time, a version-$v$ document receives ranking score
$g_v(s_v)+\epsilon s_v$, with $\epsilon=10^{-9}$. Remaining ties use the raw
score, then the original document row. This preserves the original
ordering within each version. Consequently, calibration changes only
cross-version comparisons: both homogeneous-gallery endpoints are unchanged.
The same fitted maps apply at every backfill fraction. At inference,
the maps use candidate scores and document version labels. We fit
separate maps on DEV for each method and upgrade, including methods
that convert stored vectors before re-encoding.

We split development queries by their positive image, use one half to fit
the maps, and reserve the other half for diagnostics. Each fitting query
contributes its positive and 100 negatives, split equally between hard and
uniform samples. We balance the total sample weight across datasets and
pool them into one map for each version. These maps provide ranking scales
under the sampled positive/negative ratio. The complete sampling and tie
rules are given in the supplement.

\section{Experiments}
\begin{table*}[t]\centering\small\setlength{\tabcolsep}{0pt}
\setlength{\aboverulesep}{0.65pt}\setlength{\belowrulesep}{1.15pt}
\renewcommand{\arraystretch}{0.96}
\newlength{\tablemethodwidth}\setlength{\tablemethodwidth}{1.38in}
\newlength{\tablegroupwidth}\setlength{\tablegroupwidth}{\dimexpr(\textwidth-\tablemethodwidth-27pt)/3\relax}
\newlength{\tableacolwidth}\setlength{\tableacolwidth}{.25\tablegroupwidth}
\newlength{\tablebcolwidth}
\newcolumntype{V}[1]{>{\centering\arraybackslash}m{#1}}
\newlength{\tableonescorewidth}\settowidth{\tableonescorewidth}{00.00}
\newcommand{\scorecell}[1]{\makebox[\tableonescorewidth][r]{#1}}
\makebox[\textwidth][l]{\textbf{A. Direct re-encoding}}\par\smallskip
\begin{tabular}{@{}l*{3}{@{\hspace{9pt}}*{4}{V{\tableacolwidth}}}@{}}\toprule
 & \multicolumn{4}{@{\hspace{9pt}}c}{\textbf{Qwen 2B $\rightarrow$ 2B}} & \multicolumn{4}{@{\hspace{9pt}}c}{\textbf{CLIP $\rightarrow$ Qwen 2B}} & \multicolumn{4}{@{\hspace{9pt}}c}{\textbf{Qwen 2B $\rightarrow$ 8B}}\\
 & \multicolumn{4}{@{\hspace{9pt}}c}{\footnotesize Old-system R@1 61.05} & \multicolumn{4}{@{\hspace{9pt}}c}{\footnotesize Old-system R@1 50.09} & \multicolumn{4}{@{\hspace{9pt}}c}{\footnotesize Old-system R@1 61.05}\\
\cmidrule(lr){2-5}\cmidrule(lr){6-9}\cmidrule(lr){10-13}
\multirow{2}{*}{\makebox[\tablemethodwidth][l]{Method}} & \multirow{2}{*}{\shortstack{Entry\\R@1}} & \multicolumn{2}{c}{\makebox[2\tableacolwidth]{\footnotesize Re-encoding AUC}} & \multirow{2}{*}{\shortstack{Final\\R@1}} & \multirow{2}{*}{\shortstack{Entry\\R@1}} & \multicolumn{2}{c}{\makebox[2\tableacolwidth]{\footnotesize Re-encoding AUC}} & \multirow{2}{*}{\shortstack{Final\\R@1}} & \multirow{2}{*}{\shortstack{Entry\\R@1}} & \multicolumn{2}{c}{\makebox[2\tableacolwidth]{\footnotesize Re-encoding AUC}} & \multirow{2}{*}{\shortstack{Final\\R@1}}\\
\cmidrule(lr){3-4}\cmidrule(lr){7-8}\cmidrule(lr){11-12}
 &  & {\footnotesize Native} & {\footnotesize \textbf{Calib.}} &  &  & {\footnotesize Native} & {\footnotesize \textbf{Calib.}} &  &  & {\footnotesize Native} & {\footnotesize \textbf{Calib.}} & \\\midrule
Joint\enspace{\scriptsize\color{black!75}(reference)} & \scorecell{67.45} & \scorecell{67.84} & \scorecell{69.41} & \scorecell{74.02} & \scorecell{0.06} & \scorecell{38.71} & \scorecell{37.55} & \scorecell{73.15} & \scorecell{0.03} & \scorecell{39.67} & \scorecell{38.78} & \scorecell{75.58}\\
XBT~\cite{jang2025xbt}\enspace{\scriptsize\color{black!75}ICCV '25} & \scorecell{63.76} & \scorecell{57.74} & \scorecell{67.05} & \scorecell{71.38} & \scorecell{48.47} & \scorecell{39.68} & \scorecell{58.04} & \scorecell{69.69} & \scorecell{63.84} & \scorecell{58.47} & \scorecell{67.53} & \scorecell{72.49}\\
MKP~\cite{byun2026mkp}\enspace{\scriptsize\color{black!75}ECCV '26} & \scorecell{65.07} & \scorecell{67.55} & \scorecell{68.61} & \scorecell{73.20} & \scorecell{53.03} & \scorecell{40.94} & \scorecell{59.27} & \scorecell{68.31} & \scorecell{65.51} & \scorecell{67.87} & \scorecell{69.40} & \scorecell{74.04}\\
SLERP~\cite{ricci2026slerp}\enspace{\scriptsize\color{black!75}NeurIPS '26} & \scorecell{66.71} & \scorecell{69.73} & \scorecell{69.25} & \scorecell{74.02\rlap{$^{\ddagger}$}} & \scorecell{55.83} & \scorecell{41.85} & \scorecell{61.34} & \scorecell{71.20} & \scorecell{66.82} & \scorecell{69.99} & \scorecell{70.52} & \scorecell{75.58\rlap{$^{\ddagger}$}}\\
BT$^{2*}$~\cite{zhou2023bt2}\enspace{\scriptsize\color{black!75}ICCV '23} & \scorecell{67.65} & \scorecell{63.30} & \scorecell{69.80} & \scorecell{72.69} & \scorecell{61.27} & \scorecell{42.27} & \scorecell{\textbf{65.25}} & \scorecell{72.32} & \scorecell{67.77} & \scorecell{63.22} & \scorecell{70.42} & \scorecell{74.12}\\
RACT$^{*}$~\cite{zhang2022ract}\enspace{\scriptsize\color{black!75}ICLR '22} & \scorecell{67.30} & \scorecell{\underline{70.29}} & \scorecell{\underline{70.34}} & \scorecell{74.04} & \scorecell{50.93} & \scorecell{46.13} & \scorecell{58.77} & \scorecell{69.85} & \scorecell{62.96} & \scorecell{66.66} & \scorecell{66.70} & \scorecell{71.72}\\
RankMerge~\cite{seo2025metric}\enspace{\scriptsize\color{black!75}WACV '25} & \scorecell{61.05} & \scorecell{64.85} & \scorecell{65.50} & \scorecell{74.02\rlap{$^{\ddagger}$}} & \scorecell{50.09} & \scorecell{45.13} & \scorecell{58.70} & \scorecell{73.15\rlap{$^{\ddagger}$}} & \scorecell{61.05} & \scorecell{64.85} & \scorecell{66.29} & \scorecell{75.58\rlap{$^{\ddagger}$}}\\
MCT~\cite{seo2025metric}\enspace{\scriptsize\color{black!75}WACV '25} & \scorecell{67.13} & \scorecell{69.59} & \scorecell{69.66} & \scorecell{72.67} & \scorecell{50.47} & \scorecell{\textbf{59.43}} & \scorecell{60.21} & \scorecell{71.67} & \scorecell{67.79} & \scorecell{\underline{70.49}} & \scorecell{\underline{70.56}} & \scorecell{73.80}\\
\textbf{RELAY}\enspace{\scriptsize\color{black!75}(ours)} & \scorecell{71.28} & \scorecell{\textbf{71.74}} & \scorecell{\textbf{72.23}} & \scorecell{73.46} & \scorecell{59.74} & \scorecell{\underline{49.22}} & \scorecell{\underline{62.27}} & \scorecell{72.18} & \scorecell{69.71} & \scorecell{\textbf{70.92}} & \scorecell{\textbf{71.13}} & \scorecell{75.57}\\
\bottomrule\end{tabular}\par\medskip
\makebox[\textwidth][l]{\textbf{B. Conversion-first migration}}\par\smallskip
\begingroup\footnotesize\settowidth{\tableonescorewidth}{00.00}
\setlength{\tablemethodwidth}{1.11in}
\setlength{\tablebcolwidth}{\dimexpr(\textwidth-\tablemethodwidth-27pt)/18\relax}
\begin{tabular}{@{}l*{3}{@{\hspace{9pt}}*{6}{V{\tablebcolwidth}}}@{}}\toprule
 & \multicolumn{6}{@{\hspace{9pt}}c}{\textbf{Qwen 2B $\rightarrow$ 2B}} & \multicolumn{6}{@{\hspace{9pt}}c}{\textbf{CLIP $\rightarrow$ Qwen 2B}} & \multicolumn{6}{@{\hspace{9pt}}c}{\textbf{Qwen 2B $\rightarrow$ 8B}}\\
\cmidrule(lr){2-7}\cmidrule(lr){8-13}\cmidrule(lr){14-19}
\multirow{2}{*}{\makebox[\tablemethodwidth][l]{Method}} & \multirow{2}{*}{\shortstack{Entry\\R@1}} & \multicolumn{2}{c}{\makebox[2\tablebcolwidth]{\scriptsize Conversion}} & \multicolumn{2}{c}{\makebox[2\tablebcolwidth]{\scriptsize Re-enc. AUC}} & \multirow{2}{*}{\shortstack{Final\\R@1}} & \multirow{2}{*}{\shortstack{Entry\\R@1}} & \multicolumn{2}{c}{\makebox[2\tablebcolwidth]{\scriptsize Conversion}} & \multicolumn{2}{c}{\makebox[2\tablebcolwidth]{\scriptsize Re-enc. AUC}} & \multirow{2}{*}{\shortstack{Final\\R@1}} & \multirow{2}{*}{\shortstack{Entry\\R@1}} & \multicolumn{2}{c}{\makebox[2\tablebcolwidth]{\scriptsize Conversion}} & \multicolumn{2}{c}{\makebox[2\tablebcolwidth]{\scriptsize Re-enc. AUC}} & \multirow{2}{*}{\shortstack{Final\\R@1}}\\
\cmidrule(lr){3-4}\cmidrule(lr){5-6}\cmidrule(lr){9-10}\cmidrule(lr){11-12}\cmidrule(lr){15-16}\cmidrule(lr){17-18}
 &  & {\scriptsize AUC} & {\scriptsize R@1} & {\scriptsize Native} & {\scriptsize Calib.} &  &  & {\scriptsize AUC} & {\scriptsize R@1} & {\scriptsize Native} & {\scriptsize Calib.} &  &  & {\scriptsize AUC} & {\scriptsize R@1} & {\scriptsize Native} & {\scriptsize Calib.} & \\\midrule
FastFill$^{*}$~\cite{jaeckle2023fastfill}\enspace{\scriptsize\color{black!75}ICLR '23} & \scorecell{67.45} & \scorecell{42.65} & \scorecell{67.38} & \scorecell{43.60} & \scorecell{68.55} & \scorecell{74.02\rlap{$^{\ddagger}$}} & \scorecell{0.06} & \scorecell{30.67} & \scorecell{55.46} & \scorecell{43.65} & \scorecell{62.46} & \scorecell{73.15\rlap{$^{\ddagger}$}} & \scorecell{0.03} & \scorecell{35.77} & \scorecell{66.52} & \scorecell{44.63} & \scorecell{69.35} & \scorecell{75.58\rlap{$^{\ddagger}$}}\\
$\lambda$-Orth~\cite{ricci2025lambda}\enspace{\scriptsize\color{black!75}NeurIPS '25} & \scorecell{67.36} & \scorecell{\textbf{68.18}} & \scorecell{69.95} & \scorecell{71.00} & \scorecell{71.18} & \scorecell{72.93} & \scorecell{52.14} & \scorecell{\textbf{36.74}} & \scorecell{59.12} & \scorecell{\underline{62.90}} & \scorecell{\underline{65.02}} & \scorecell{71.80} & \scorecell{66.68} & \scorecell{\textbf{67.48}} & \scorecell{71.11} & \scorecell{\underline{72.57}} & \scorecell{\underline{72.64}} & \scorecell{74.86}\\
BiCT$^{*\dagger}$~\cite{su2022bict}\enspace{\scriptsize\color{black!75}arXiv '22} & \scorecell{63.16} & \scorecell{66.15} & \scorecell{70.17} & \scorecell{\underline{71.48}} & \scorecell{\underline{71.74}} & \scorecell{73.56} & \scorecell{33.60} & \scorecell{32.63} & \scorecell{53.64} & \scorecell{61.59} & \scorecell{61.80} & \scorecell{71.81} & \scorecell{42.36} & \scorecell{\underline{42.38}} & \scorecell{65.46} & \scorecell{68.30} & \scorecell{68.48} & \scorecell{71.80}\\
EC$^{\dagger}$~\cite{yoon2025converter}\enspace{\scriptsize\color{black!75}ACL '25} & \scorecell{67.45} & \scorecell{\underline{66.66}} & \scorecell{72.07} & \scorecell{\textbf{72.96}} & \scorecell{\textbf{72.91}} & \scorecell{72.07} & \scorecell{0.06} & \scorecell{\underline{33.21}} & \scorecell{61.00} & \scorecell{\textbf{67.00}} & \scorecell{\textbf{66.59}} & \scorecell{61.00} & \scorecell{0.03} & \scorecell{38.24} & \scorecell{72.13} & \scorecell{\textbf{73.71}} & \scorecell{\textbf{73.59}} & \scorecell{72.13}\\
\bottomrule\end{tabular}\endgroup
\caption{\textbf{Retrieval quality during model upgrades.} Macro text-to-image Recall@1 (\%, TEST). MKP denotes MKP-Adapter, RankMerge denotes RankMerge-basic, and EC denotes Embedding-Converter. Entry uses the original gallery in both panels. For SLERP, DEV Entry R@1 selects $\alpha=1$ (SVD alignment) in both Qwen-to-Qwen upgrades and $\alpha=0.5$ in CLIP-to-Qwen. Re-encoding starts from the original gallery in A and the fully converted gallery in B. Conversion R@1 is measured at full conversion. Calib. applies DEV-fitted isotonic maps to the same features as Native. Bold and underline mark the top two in each AUC column. $^{*}$ Classification objectives adapted to paired retrieval. $^{\dagger}$ Re-encoding extends BiCT and EC. $^{\ddagger}$ Final ranking equals Joint's by construction. EC's Final R@1 follows full conversion. After query refinement, RELAY (P100) trails Joint by 0.56, 0.97, and 0.01 percentage points in Final R@1, respectively.}
\label{tab:projection-main}
\end{table*}

\begin{figure*}[t]\centering
\includegraphics[width=\textwidth]{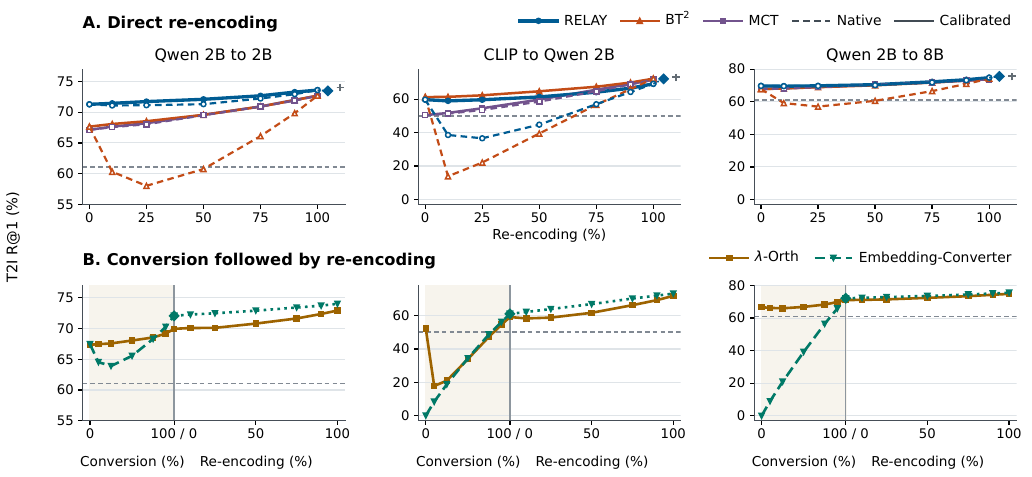}
\caption{\textbf{Retrieval during gallery migration.}
(A) Direct re-encoding starts on the original gallery. Dashed lines
show native scores and solid lines use matched isotonic calibration.
The diamond (RELAY after refinement) and plus (Joint final R@1) fall
outside the AUC.
(B) The conversion and re-encoding axes meet at 100/0\%; their widths
are for display. Curves use native scores. The diamond marks
Embedding-Converter's endpoint and the dotted line its re-encoding
extension. Gray dashed lines mark the old system. Markers show all seven
fractions per phase, and both rows share the R@1 scale within each setting.}
\label{fig:curves}\end{figure*}
\subsection{Setup}
\paragraph{Data and upgrades.}
We use paired captions and images from COCO and Flickr30K~\cite{lin2014coco,young2014flickr}.
COCO uses a deterministic 5,000-image holdout from train2017 for the
development set and all 5,000 val2017 images for the test set (TEST). Flickr30K
uses the released Karpathy split~\cite{karpathy2015alignments} with 1,014
development and 1,000 test images. \Cref{sec:data-splits} gives the training
pools, overlap checks, and sampling schedule.
The source models are Qwen3-VL-Embedding 2B~\cite{li2026qwenvlembedding}
and CLIP ViT-L/14~\cite{radford2021clip}; targets are Qwen3-VL-Embedding
2B and 8B. The shared retrieval dimensions are 2048 for Qwen-to-Qwen and
768 for CLIP-to-Qwen. BT$^2$ adds 32 dimensions in each setting.

\paragraph{Training and selection.}
We train independent role LoRAs (rank 32, alpha 64, dropout 0.05) on eight
GPUs with 128 images and 640 captions per step. RELAY uses 2880/960/960
steps for query adaptation, document adaptation, and query refinement,
with temperature $\tau=0.02$. We weight the prefixes
$\mathcal K=\{2048,1024,768,512\}$ equally for Qwen-to-Qwen and use
$\mathcal K=\{768\}$ for CLIP-to-Qwen. Qwen query adaptation uses the first
2880 steps of a 5760-step cosine schedule at LR $10^{-4}$. CLIP query
adaptation uses a full 2880-step schedule at $3\times10^{-4}$. The later
stages use $5\times10^{-4}$. Model selection and score calibration use
the development set (\cref{sec:development-selection}).

\paragraph{Evaluation and baselines.}
We evaluate single upgrades by exact text-to-image retrieval, with
Recall@1 averaged equally across the two datasets. All methods use the
same galleries, relevance labels, nested replacement order, and seven
fractions $\{0,0.1,0.25,0.5,0.75,0.9,1\}$. \Cref{tab:projection-main} uses
one run per method and upgrade, with a common replacement order.
We fit method-specific isotonic maps under the same DEV protocol and
evaluate them on fixed features. Final query refinement is measured
separately from the backfill curve.

\paragraph{Choice of Joint reference.}
Joint and RELAY share the target backbone, independent rank-32 role
adapters, retrieval loss, training examples, and global batch. Joint
updates the query and document roles together. For each upgrade to 2B,
we use a single learning-rate schedule for the 2880-step run and
compare single and staged schedules at 4800 steps. The 2880-step recipe
gives the highest development-set Recall@1 in both settings
(\cref{tab:joint-dev-main}). We train the
8B Joint reference for 2880 steps with fresh role adapters over the
8B backbone. The schedules and test endpoints are in
\cref{tab:joint-selection,sec:capacity-baselines}.

\begin{table}[t]\centering\small
\begin{tabular}{lrr}\toprule
Joint recipe & \shortstack{2B$\to$2B\\2048-d eval.} &
\shortstack{CLIP$\to$2B\\768-d eval.} \\\midrule
2880 steps, single schedule & \textbf{74.54} & \textbf{73.71} \\
4800 steps, single schedule & 73.93 & 73.24 \\
4800 steps, staged schedule & 74.23 & 73.14 \\
\bottomrule\end{tabular}
\caption{Joint schedule selection for the 2B target. The query and document
encoders use Qwen 2B in each setting. For 2B-to-2B, training averages losses over the
2048-, 1024-, 768-, and 512-dimensional prefixes. For CLIP-to-2B, training
uses the 768-dimensional prefix. We choose the recipe with the highest
development-set Recall@1 (\%), averaged equally across COCO and Flickr30K.}
\label{tab:joint-dev-main}
\end{table}

The selected Joint run performs 2880 updates per role. RELAY performs
2880 query updates and 960 document updates through backfill, with a
further 960 query updates for the refined endpoint. The comparison
therefore measures the quality of the two training recipes under a
shared retrieval setup, with their respective allocations of role
updates. \Cref{tab:projection-cost} reports their measured training cost.

\paragraph{Baseline training.}
Mapping baselines use the corresponding rank-32 Joint teacher where
applicable. RACT, FastFill, BT$^2$, and BiCT adapt classification
objectives to paired retrieval. Embedding-Converter fits a map between
frozen document features. These methods use their own training and
scoring procedures within the same retrieval evaluation. The supplement
details each adaptation and its selection budget.

\subsection{Access, migration, and final quality}

\paragraph{Entry on the untouched gallery.}
Deploying the new query encoder alone improves R@1 by 10.23, 9.65, and
8.66 points over the original systems (\cref{tab:projection-main}).
Across direct re-encoding and conversion-first methods, RELAY ranks first
in the two Qwen-to-Qwen upgrades at 71.28\% and 69.71\%. In CLIP-to-Qwen,
RELAY ranks second at 59.74\%, behind BT$^2$ at 61.27\%.

\paragraph{Re-encoding with and without prior conversion.}
RELAY starts re-encoding from the original gallery. Among direct
methods, it ranks first in both Qwen-to-Qwen upgrades and second in
CLIP-to-Qwen under both scoring protocols. In CLIP-to-Qwen, RELAY
reaches 49.22\% native AUC behind MCT (59.43\%) and 62.27\% calibrated
AUC behind BT$^2$ (65.25\%) (\cref{tab:projection-main}A).
Conversion-first methods apply a lightweight map to the stored vectors
before re-encoding the source images. Re-encoding therefore begins on a
fully converted gallery. In CLIP-to-Qwen, the re-encoding extension of
Embedding-Converter starts that phase at 61.00\% R@1 and reaches 67.00\%
native AUC (\cref{tab:projection-main}B and \cref{fig:curves}B).
That AUC covers the phase after conversion. \Cref{sec:deployment-work}
compares the gallery work required by the two paths.

\paragraph{Sensitivity to replacement order and training.}
The calibrated AUC rankings against the strongest direct baselines hold
across six replacement orders (\cref{sec:replacement-order}). Across three
training runs, RELAY leads RACT by 1.87--1.90 points in calibrated AUC for
2B-to-2B and MCT by 0.20--0.56 points for 2B-to-8B. In the latter upgrade,
MCT leads on native scores in one run (\cref{tab:paired-training-repeats}).

\paragraph{Final retrieval.}
After query refinement, RELAY finishes within 0.97 points of Joint
across the three upgrades (\cref{tab:projection-main}A). Refinement
changes the query encoder alone. It lifts CLIP-to-Qwen from 69.22\% to
72.18\% and lowers 2B-to-2B from 73.59\% to 73.46\%.
RankMerge-basic and FastFill recover Joint after full re-encoding.
Pure SVD preserves Joint's ranking in the two Qwen settings.
Embedding-Converter ends at full conversion.

\section{Analysis}
\subsection{Where to use a learned projection}
\begin{table*}[t]\centering\small\setlength{\tabcolsep}{5pt}
\begin{tabular}{llrrrrrcc}\toprule
\multirow{2}{*}{Setting} & \multirow{2}{*}{Mask} & \multirow{2}{*}{Start} & \multirow{2}{*}{Raw AUC} & \multirow{2}{*}{Calibrated AUC} & \multirow{2}{*}{\shortstack{Backfill\\end}} & \multirow{2}{*}{\shortstack{Final\\system}} & \multicolumn{2}{c}{Final projections}\\
\cmidrule(lr){8-9}
 & & & & & & & Q & D\\\midrule
2B-to-2B & P000 & 70.81 & 71.11 & 72.00 & \textbf{73.62} & \textbf{73.85} & $\times$ & $\times$\\
 & P111 & \textbf{71.28} & 64.21 & 70.30 & 72.69 & 72.04 & $\checkmark$ & $\checkmark$\\
 & P110 & \textbf{71.28} & 64.21 & 70.30 & 72.69 & 73.25 & $\times$ & $\checkmark$\\
 & P100 & \textbf{71.28} & \textbf{71.74} & \textbf{72.23} & 73.59 & 73.46 & $\times$ & $\times$\\
\midrule
CLIP-to-2B & P000 & 54.58 & 40.16 & 58.14 & 66.84 & 70.64 & $\times$ & $\times$\\
 & P111 & \textbf{59.74} & 43.39 & 61.47 & \textbf{70.44} & 71.16 & $\checkmark$ & $\checkmark$\\
 & P110 & \textbf{59.74} & 43.39 & 61.47 & \textbf{70.44} & 71.72 & $\times$ & $\checkmark$\\
 & P100 & \textbf{59.74} & \textbf{49.22} & \textbf{62.27} & 69.22 & \textbf{72.18} & $\times$ & $\times$\\
\midrule
2B-to-8B & P000 & 60.46 & 45.05 & 64.17 & 70.90 & 74.21 & $\times$ & $\times$\\
 & P111 & \textbf{69.71} & 59.09 & 66.64 & 73.54 & 73.08 & $\checkmark$ & $\checkmark$\\
 & P110 & \textbf{69.71} & 59.09 & 66.64 & 73.54 & 74.44 & $\times$ & $\checkmark$\\
 & P100 & \textbf{69.71} & \textbf{70.92} & \textbf{71.13} & \textbf{74.77} & \textbf{75.57} & $\times$ & $\times$\\
\bottomrule\end{tabular}
\caption{Stage projection ablations (\%, TEST). Bits specify a learned output projection on the trained role at each stage. P100 and P111 share the entry query. P110 and P111 share the backfill curve. Final Q/D marks indicate retained ($\checkmark$) or absent ($\times$) learned output projections. The adapted LoRAs remain in both cases. Bold marks best displayed values within each setting.}\label{tab:projection-ablation}
\end{table*}

\begin{table*}[t]\centering\small\setlength{\tabcolsep}{5pt}
\begin{tabular}{llrrrrrr}\toprule Setting & Recipe & Stage 1 & Stage 2 & Stage 3 & Sum & Joint & Reduction\\\midrule
2B-to-2B & P000 & 11.32 & 4.10 & 6.63 & 22.06 & 25.35 & 13.0\%\\
2B-to-2B & P100 & 10.78 & 4.06 & 6.63 & 21.47 & 25.35 & 15.3\%\\
CLIP-to-2B & P000 & 11.10 & 2.78 & 3.70 & 17.58 & 25.32 & 30.6\%\\
CLIP-to-2B & P100 & 11.32 & 2.78 & 3.69 & 17.79 & 25.32 & 29.7\%\\
2B-to-8B & P000 & 15.94 & 9.59 & 18.69 & 44.22 & 67.22 & 34.2\%\\
2B-to-8B & P100 & 16.00 & 9.55 & 18.80 & 44.36 & 67.22 & 34.0\%\\
\bottomrule\end{tabular}
\caption{Training GPU-hours. The compared runs use eight GPUs of the same model. RELAY includes all three stages and counts reused query adaptation once per recipe. Joint trains both roles for 2880 steps. Totals include loading and saving within successful training processes. The supplement describes the scope of feature-encoding measurements.}\label{tab:projection-cost}
\end{table*}

The three bits in P000, P111, P100, and P110 indicate whether the role
trained at each stage uses a learned output projection. The frozen role
carries over its complete function from the previous stage.

\paragraph{Effect of the query projection.}
P000 trains LoRA with fixed prefixes at all stages. P100 adds a learned query projection to
$Q_1$ and keeps it through backfill. Both use fixed prefixes for $D_1$ and
the refined query. Each run trains $D_1$ against its own $Q_1$. In 2B-to-2B, P100 gains
0.47 points at entry and 0.63 points in raw backfill AUC, while P000
finishes slightly higher (73.85\% versus 73.46\%). These small gaps are
consistent with the shared pretrained backbone providing enough initial
alignment for LoRA to adapt the query. In CLIP-to-2B and 2B-to-8B,
P100 gains 5.16 and 9.25 points at entry and 9.06 and 25.87 points in
raw backfill AUC, respectively (\cref{tab:projection-ablation}). In these
two upgrades, the projection can combine coordinates from the full
target embedding, while P000 uses its fixed prefix.

\paragraph{Effect of the document projection.}
P100 and P110 differ only in the projection setting for document
adaptation. Both train a rank-32 document LoRA against the same frozen
$Q_1$, including its learned query projection, with the same data and
schedule. P100 retains the first $d$ coordinates of the document embedding,
and P110 learns a document projection to $d$ dimensions. The fixed prefix
yields higher raw backfill AUC by 7.53, 5.82, and 11.82 points across the three settings
(\cref{tab:projection-ablation}). The gains occur on both datasets
(\cref{tab:projection-pools}).
This comparison tests the document-side design, including the change
in parameter count, with the deployed query held fixed.

In CLIP-to-Qwen, the fixed document prefix gives higher backfill AUC
than the learned document projection under both native scoring
(49.22\% vs.\ 43.39\%) and matched calibration (62.27\% vs.\ 61.47\%).
At full backfill, the learned projection gives higher Recall@1 with
the same $Q_1$ (70.44\% vs.\ 69.22\%).

\paragraph{Removing the query projection.}
P110 and P111 share the same backfill curve. P110 removes the query
projection and refines the query against the same fixed documents,
improving final Recall@1 by 1.21, 0.56, and 1.36 points. In CLIP-to-Qwen,
P100 reaches 69.22\% at full backfill. Removing $W_Q$ lowers Recall@1
to 67.39\% before further training. Query refinement then raises it
to 72.18\% using the same document vectors. In 2B-to-8B, Recall@1 drops
from 74.77\% to 73.40\% after projection removal, then rises to 75.57\%
after refinement. The backfill endpoint also supports serving with $Q_1$.

\paragraph{Training documents against the deployed query.}
We fix $Q_1$, $D_0$, and the replacement order, and compare $D_1$ with
Joint document vectors linearly mapped to the $D_0$ space.
\Cref{tab:fixed-query-control} shows higher calibrated AUC and higher
end-of-backfill R@1 for RELAY in all three upgrades. In CLIP-to-Qwen,
RELAY trails by 11.50 points in native AUC and leads by 1.01 after
calibration, with a 0.75-point lower sampled minimum.
\Cref{sec:fixed-query-control} details the fit and curves.
\begin{table}[t]\centering\small\setlength{\tabcolsep}{3pt}
\begin{tabular}{lrrrr}\toprule
Upgrade & \shortstack{Native\\AUC} & \shortstack{Calib.\\AUC} & \shortstack{Calib.\\min.} & \shortstack{100\%\\R@1} \\
\midrule
2B-to-2B & +0.50 & +0.41 & 0.00 & +0.94 \\
CLIP-to-2B & -11.50 & +1.01 & -0.75 & +5.06 \\
2B-to-8B & +1.05 & +0.97 & -0.17 & +3.07 \\
\bottomrule\end{tabular}
\caption{\textbf{Fixed-query document control.} Differences (RELAY
minus mapped Joint, points) on TEST. Both arms share $Q_1$ throughout.
The minimum is taken over the seven fractions.}
\label{tab:fixed-query-control}
\end{table}

\subsection{Score calibration in mixed galleries}
\begin{table}[t]\centering\small\setlength{\tabcolsep}{4pt}
\begin{tabular}{lrrrr}\toprule
Setting & Start & \shortstack{Backfill\\end} & \shortstack{Raw\\min.} & \shortstack{Calibrated\\min.}\\\midrule
2B-to-2B & 71.28 & 73.59 & 71.09 & 71.28\\
CLIP-to-2B & 59.74 & 69.22 & 36.60 & 58.99\\
2B-to-8B & 69.71 & 74.77 & 69.58 & 69.53\\
\bottomrule\end{tabular}
\caption{P100 endpoint and mixed-gallery diagnostics, TEST macro R@1 (\%). Minima use the seven measured fractions. The same features are used with and without calibration; Start and Backfill end are unchanged. Final query refinement is excluded.}\label{tab:mixing-diagnostic}
\end{table}

\paragraph{Across methods and deployment paths.}
We hold each method's encoders, embeddings, and replacement order fixed
and fit its score maps on DEV. Across \cref{tab:projection-main},
calibration raises re-encoding AUC in 33 of 39 method--upgrade pairs.
AUC improves in 24 of 27 direct cases and 9 of 12 conversion-first
cases. It decreases in six cases. In CLIP-to-Qwen, BT$^2$ improves by
22.98 points during direct re-encoding and FastFill by 18.81 points
during re-encoding after conversion.

\paragraph{Effect on RELAY.}
Calibration raises AUC by 0.49 points in 2B-to-2B, 13.05 points in
CLIP-to-Qwen, and 0.21 points in 2B-to-8B. In CLIP-to-Qwen the sampled
minimum rises from 36.60\% to 58.99\%, above the original system's
50.09\%. In 2B-to-8B the sampled minimum moves from 69.58\% to 69.53\%.
Across the six CLIP-to-Qwen orders, finer sampling through 25\% backfill
finds native minima of 35.33--36.23\% at 17.5--20\% backfill. The minima
over the same early interval rise to 58.97--59.45\% with calibration
(\cref{fig:clip-early-backfill}). Per-dataset AUCs are in
\cref{tab:calibration-pool-summary}.

For a caption with one relevant image, a top-1 hit requires the
image to rank first among documents of its own version and then to
outrank the best candidate from the other version. The monotone maps
preserve the first comparison and can change the second. This accounts
for RELAY's fixed entry and backfill endpoints under both score protocols and
explains how the mixed-gallery minimum can change substantially while
the underlying representations are held fixed.

\subsection{Measured training resources}
The compared runs use eight GPUs of the same model and an effective batch
size of 128 images. P100 uses 15.3\%, 29.7\%, and 34.0\% fewer training GPU-hours than Joint
(\cref{tab:projection-cost}). The totals sum the observed training
processes for each recipe, including loading and saving. Joint updates
both roles at each step. RELAY uses a frozen counterpart for each stage,
evaluated online in the later Qwen stages and read from cached features
in the later CLIP stages. We compute GPU-hours by multiplying the elapsed
hours of each successful training process by its number of GPUs and
summing over the selected stages. Although RELAY has more optimizer
steps in total, each step updates one role. The measured differences
reflect the full recipes, including their schedules and use of cached
features. The supplement describes the scope of the separate
feature-encoding measurements.

\subsection{Gallery work during deployment}
\label{sec:deployment-work}
For a gallery of $N$ documents, RELAY can serve with $Q_1$ while all
$N$ old vectors are still in place. Its direct path encodes each source
document once with $D_1$ and replaces that document's active vector.
The optional switch to the refined query reuses the completed gallery.
This path assumes access to the original documents. With a forward
converter, the first pass can run on stored vectors alone. A method
that converts the full gallery and then re-encodes its documents makes
one $N$-item pass over the vectors and one $N$-item pass over the source
documents. Embedding-Converter can also stop after the first pass.

In CLIP-to-Qwen, RELAY starts re-encoding on the original gallery
at 59.74\% Recall@1. $\lambda$-Orth reaches 59.12\% after converting
all stored vectors and starts re-encoding from that gallery
(\cref{tab:projection-main}). Each phase's AUC tracks recall as a function
of the fraction of updates completed. Comparing elapsed time would
also require measuring vector conversion, document encoding, and index updates.

The calibrated path uses one $Q_1$ vector for both versions and a
version tag to select the score map. An exact implementation can take
the raw top $K$ under each version filter, calibrate those candidates,
and merge at most $2K$ results. Every omitted document has at least
$K$ documents from its own version ahead of it under either scoring
rule. With the same tie rule, the merge recovers the exact calibrated
top $K$. A deployed approximate index would
also need to account for candidate recall and the cost of filtering.

\Needspace{10\baselineskip}
\section{Conclusion}
Training the query encoder first lets RELAY improve retrieval before
backfill starts and use that encoder throughout a single re-encoding pass.
RELAY ranks in the top two by direct re-encoding AUC in all three upgrades
under both native and matched calibrated scoring. Across both deployment
paths, per-version score calibration improves re-encoding AUC in 33 of 39
method--upgrade pairs on fixed embeddings. After query refinement,
RELAY finishes within 0.97 percentage points of Joint in Recall@1
while keeping the re-encoded gallery in place.

\label{mainend}
\Needspace{6\baselineskip}
\section*{LLM Usage Statement}
OpenAI Codex and Qwen assisted with implementing training and evaluation
workflows, checking experimental results, and drafting and revising the
manuscript.
The authors take full responsibility for the paper's content and conclusions.
{\small\setlength{\bibsep}{0pt}\linespread{0.96}\selectfont
\bibliographystyle{ieeenat_fullname}\bibliography{references}}
\clearpage\appendix\onecolumn\raggedbottom
\section{Supplementary Material}
We provide data construction, implementation details, baseline adaptations,
and evaluation results for the three upgrades. All reported experiments use
the public COCO/Flickr training mixture.

\subsection{Current three-scenario projection experiments}

\subsubsection{Parameterization, initialization and frozen functions}
The two target roles have independent rank 32/alpha 64/dropout 0.05 LoRAs over
frozen Qwen backbones. A learned output projection is linear, with no bias
or activation, and FP32 weights initialized to identity rows. The matrices are
$2048\times2048$ for same-base, $768\times2048$ for CLIP transfer, and
$2048\times4096$ for capacity upgrade. Projection reads the full native
embedding before the prefix loss. Roles without a projection use a fixed
prefix. The two Qwen-to-Qwen settings use equally weighted
2048/1024/768/512 retrieval prefixes; CLIP uses 768. The retrieval temperature
for RELAY is 0.02 in all three stages. These choices retain the original paired contrastive
objective, with no extra compatibility or classification loss.

P100 branches from the exact P111 entry checkpoint. During document
adaptation, the frozen $Q_1$ retains its LoRA and query projection. The
document role starts from its own target initialization and uses a fixed
prefix. Query refinement starts with the adapted query LoRA weights and
removes the query projection before training against the fixed $D_1$.
P110 branches from P111 after document adaptation and keeps the document
projection through query refinement. The final P110 pair retains only
the document projection, P111 retains both projections, and P000 and
P100 use fixed prefixes on both sides.

Each new stage starts a fresh optimizer. The common optimizer is fused
AdamW, betas $(0.9,0.95)$, epsilon $10^{-8}$, weight decay 0.1, clipping 1, BF16
backbone/LoRA, nominal GradCache replay chunk 16, and cosine warmup ratio 0.001.
All stages adopt 2880/960/960 steps; the Qwen stage 1 schedule has horizon 5760 at
LR $10^{-4}$, while CLIP stage 1 has horizon 2880 at $3\times10^{-4}$. Stages 2 and 3 use
LR $5\times10^{-4}$ and horizon 960. World size 8, 16 image groups per rank and
accumulation 1 give 128 images/640 captions per optimizer step. One pass
contains 122,880 logical draws; deduplicated physical-image encoding may
have 122,856 rows with aliases preserving all logical examples.

The Qwen stage 1 teacher is the accepted official 2B Doc TRAIN cache. Qwen stages 2 and 3
evaluate the appropriate fixed role online in evaluation/no-gradient mode.
CLIP stages use frozen-feature banks: stage 2 needs its own $Q_1$ TRAIN
features and stage 3 needs its own $D_1$ TRAIN features. Frozen-side numerical
boundaries are preserved: Qwen's genuine SDK-derived FP32-L2 teacher is
cast directly to BF16; CLIP's accepted F16-derived feature path is kept,
including the FP32 projection where present before the training-view cast.
Evaluation uses the accepted FP32-L2 views and exact FP64 dot scoring.
These cache and dtype paths are why current matched P000 results should
not be silently substituted for an older online/cached LoRA run.

\subsubsection{Capacity-upgrade baselines}
\label{sec:capacity-baselines}
Capacity uses a true independent 8B Joint2880 teacher, with both roles
trained at LR $5\times10^{-4}$ over its full 2880-step schedule. Its native
query is evaluated at all seven fractions against the nested old/new gallery,
giving TEST entry 0.03\%, AUC 39.67\%, and endpoint 75.58\%. MCT uses its complete
$\rho+\psi+$native MCL method with paired T2I. DEV selects epoch 50
by native re-encoding AUC for seed 42.
$\lambda$-Orth uses the paper objective and its fixed 100-epoch endpoint,
with conversion and re-encoding both evaluated. SLERP uses $\alpha=1$,
selected by DEV Entry R@1. RACT uses its fixed same-instance setting
$\lambda=1$. We evaluate these baselines with their native scores.
Shared representation dimensions are 2048 for capacity and
same-base, and 768 for CLIP.

\paragraph{BT$^2$, BiCT, and Embedding-Converter.}
BT$^2$ uses a 2880-step student with a learned basis transformation and
32 additional coordinates. Paired retrieval replaces the classification
objectives, with cosine supervision from the frozen old and Joint teachers.
The mixed-gallery evaluation zero-pads old documents and scores all vectors
with the expanded query. This evaluation reports retrieval quality under
that dimension choice; index costs are outside its measurement scope.
BiCT uses 2880 encoder steps with paired retrieval and a same-input cosine
compatibility term at $\lambda=1$, followed by 900 steps of cosine regression
from old document features to the compatible new document features.
Embedding-Converter freezes both models and fits the forward converter on
paired TRAIN document vectors using absolute-error regression and global and
local distance preservation. Converted-gallery DEV R@1 selects a
checkpoint from a 50,000-step fit. The fully converted gallery is the
Embedding-Converter endpoint. We evaluate an additional re-encoding
phase for BiCT and Embedding-Converter with the same replacement order.

\subsubsection{Complete stage curves and endpoint interventions}
\begin{center}
\includegraphics[width=0.98\linewidth]{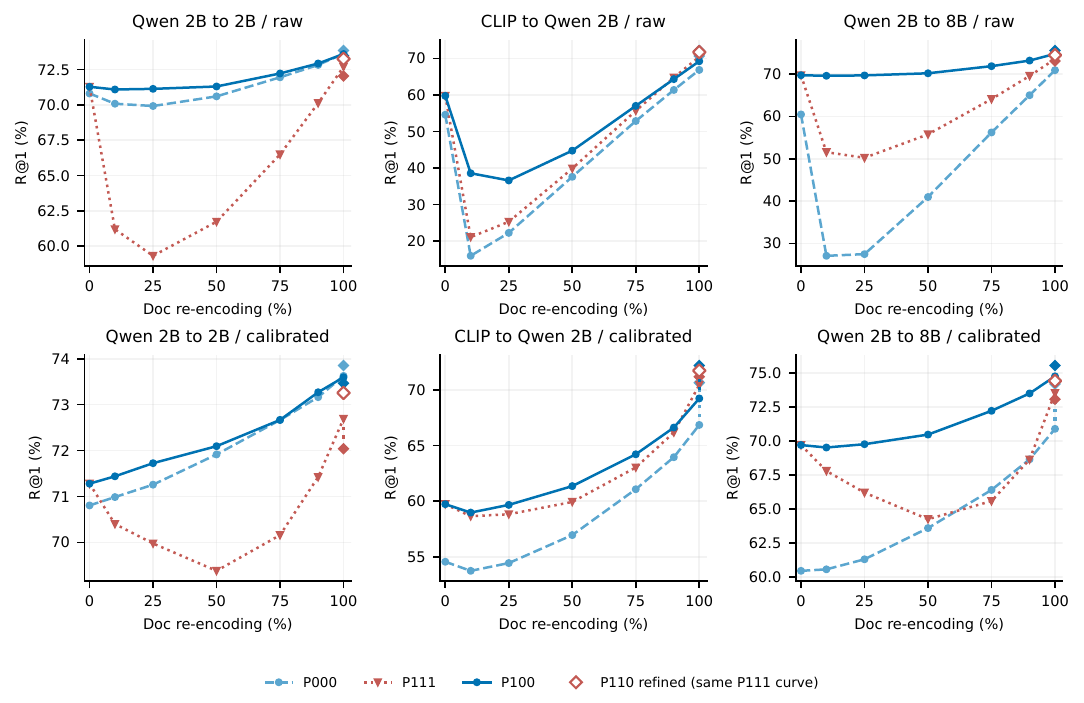}
\captionof{figure}{\textbf{Stage masks under raw and calibrated scoring.}
Each line is a measured fixed-query curve. P110 inherits P111's curve;
its distinct refined endpoint is the hollow diamond. Other diamonds mark the respective
refined endpoints. All points use seed 42, and query refinement is excluded from AUC.}
\label{fig:projection-masks}\end{center}

\begingroup\small\setlength{\LTpre}{0pt}\setlength{\LTpost}{3pt}
\Needspace{24\baselineskip}
\begin{longtable}{lrrrrrrr}
\caption{Qwen 2B to 2B: seed 42 TEST macro R@1 (\%) at every measured fraction.}\label{tab:fractions-same2048}\\
\toprule
Method / phase & 0\% &10\%&25\%&50\%&75\%&90\%&100\%\\\midrule
\endfirsthead
\toprule Method / phase &0\%&10\%&25\%&50\%&75\%&90\%&100\%\\\midrule\endhead
Joint / re-enc. & 67.45 & 64.96 & 64.75 & 66.82 & 69.89 & 72.07 & 74.02\\
XBT / re-enc. & 63.76 & 53.11 & 50.17 & 54.14 & 61.83 & 67.29 & 71.38\\
MKP / re-enc. & 65.07 & 64.49 & 64.84 & 66.90 & 69.64 & 71.66 & 73.20\\
SLERP / re-enc. & 66.71 & 67.01 & 67.69 & 69.54 & 71.41 & 72.89 & 74.02\\
BT$^{2*}$ / re-enc. & 67.65 & 60.31 & 58.02 & 60.75 & 66.10 & 69.83 & 72.69\\
RACT$^{*}$ / re-enc. & 67.30 & 67.78 & 68.40 & 70.29 & 71.86 & 73.01 & 74.04\\
RankMerge / re-enc. & 61.05 & 61.42 & 62.07 & 63.77 & 66.70 & 69.76 & 74.02\\
MCT / re-enc. & 67.13 & 67.56 & 68.03 & 69.50 & 70.92 & 71.88 & 72.67\\
FastFill$^{*}$ / conv. & 67.45 & 61.73 & 52.91 & 38.57 & 24.82 & 24.00 & 67.38\\
FastFill$^{*}$ / re-enc. & 67.38 & 15.98 & 22.67 & 40.19 & 57.58 & 67.44 & 74.02\\
$\lambda$-Orth / conv. & 67.36 & 67.45 & 67.57 & 68.05 & 68.54 & 69.17 & 69.95\\
$\lambda$-Orth / re-enc. & 69.95 & 70.08 & 70.11 & 70.80 & 71.64 & 72.37 & 72.93\\
BiCT$^{*\dagger}$ / conv. & 63.16 & 63.35 & 63.91 & 66.07 & 68.10 & 69.22 & 70.17\\
BiCT$^{*\dagger}$ / re-enc. & 70.17 & 70.41 & 70.73 & 71.26 & 72.08 & 72.91 & 73.56\\
EC$^{\dagger}$ / conv. & 67.45 & 64.54 & 63.90 & 65.55 & 68.30 & 70.28 & 72.07\\
EC$^{\dagger}$ / re-enc. & 72.07 & 72.28 & 72.48 & 72.90 & 73.42 & 73.72 & 74.02\\
RELAY--P000 / re-enc. & 70.81 & 70.08 & 69.92 & 70.60 & 71.95 & 72.83 & 73.62\\
RELAY--P000 + Calibration / re-enc. & 70.81 & 70.99 & 71.26 & 71.92 & 72.67 & 73.16 & 73.62\\
RELAY--P111 / re-enc. & 71.28 & 61.20 & 59.32 & 61.73 & 66.49 & 70.12 & 72.69\\
RELAY--P111 + Calibration / re-enc. & 71.28 & 70.40 & 69.98 & 69.38 & 70.16 & 71.42 & 72.69\\
RELAY--P100 / re-enc. & 71.28 & 71.09 & 71.13 & 71.30 & 72.22 & 72.93 & 73.59\\
RELAY--P100 + Calibration / re-enc. & 71.28 & 71.44 & 71.73 & 72.10 & 72.67 & 73.27 & 73.59\\
\bottomrule\end{longtable}
\Needspace{24\baselineskip}
\begin{longtable}{lrrrrrrr}
\caption{CLIP to Qwen 2B: seed 42 TEST macro R@1 (\%) at every measured fraction.}\label{tab:fractions-clip768}\\
\toprule
Method / phase & 0\% &10\%&25\%&50\%&75\%&90\%&100\%\\\midrule
\endfirsthead
\toprule Method / phase &0\%&10\%&25\%&50\%&75\%&90\%&100\%\\\midrule\endhead
Joint / re-enc. & 0.06 & 9.03 & 21.24 & 39.65 & 57.06 & 66.81 & 73.15\\
XBT / re-enc. & 48.47 & 10.02 & 20.70 & 38.17 & 54.40 & 63.65 & 69.69\\
MKP / re-enc. & 53.03 & 17.66 & 22.77 & 38.01 & 53.86 & 62.69 & 68.31\\
SLERP / re-enc. & 55.83 & 15.25 & 22.41 & 39.20 & 55.82 & 65.16 & 71.20\\
BT$^{2*}$ / re-enc. & 61.27 & 13.81 & 22.07 & 39.46 & 56.64 & 66.05 & 72.32\\
RACT$^{*}$ / re-enc. & 50.93 & 32.13 & 31.47 & 42.11 & 56.04 & 64.39 & 69.85\\
RankMerge / re-enc. & 50.09 & 24.11 & 27.63 & 41.92 & 57.83 & 67.06 & 73.15\\
MCT / re-enc. & 50.47 & 51.32 & 53.56 & 58.55 & 64.74 & 68.89 & 71.67\\
FastFill$^{*}$ / conv. & 0.06 & 8.04 & 18.26 & 31.88 & 44.38 & 51.11 & 55.46\\
FastFill$^{*}$ / re-enc. & 55.46 & 19.81 & 24.29 & 40.36 & 57.20 & 66.84 & 73.15\\
$\lambda$-Orth / conv. & 52.14 & 17.79 & 21.02 & 33.93 & 47.23 & 54.56 & 59.12\\
$\lambda$-Orth / re-enc. & 59.12 & 58.28 & 58.77 & 61.65 & 66.22 & 69.24 & 71.80\\
BiCT$^{*\dagger}$ / conv. & 33.60 & 13.40 & 19.41 & 31.52 & 43.23 & 49.48 & 53.64\\
BiCT$^{*\dagger}$ / re-enc. & 53.64 & 54.27 & 56.33 & 61.13 & 66.51 & 69.44 & 71.81\\
EC$^{\dagger}$ / conv. & 0.06 & 8.36 & 18.80 & 34.39 & 48.66 & 56.20 & 61.00\\
EC$^{\dagger}$ / re-enc. & 61.00 & 62.14 & 63.90 & 66.86 & 70.19 & 71.95 & 73.15\\
RELAY--P000 / re-enc. & 54.58 & 15.97 & 22.25 & 37.59 & 52.86 & 61.36 & 66.84\\
RELAY--P000 + Calibration / re-enc. & 54.58 & 53.77 & 54.46 & 56.97 & 61.08 & 63.95 & 66.84\\
RELAY--P111 / re-enc. & 59.74 & 21.09 & 25.30 & 39.85 & 55.66 & 64.75 & 70.44\\
RELAY--P111 + Calibration / re-enc. & 59.74 & 58.66 & 58.84 & 59.94 & 63.03 & 66.17 & 70.44\\
RELAY--P100 / re-enc. & 59.74 & 38.57 & 36.60 & 44.75 & 57.00 & 64.33 & 69.22\\
RELAY--P100 + Calibration / re-enc. & 59.74 & 58.99 & 59.68 & 61.36 & 64.21 & 66.61 & 69.22\\
\bottomrule\end{longtable}
\Needspace{24\baselineskip}
\begin{longtable}{lrrrrrrr}
\caption{Qwen 2B to 8B: seed 42 TEST macro R@1 (\%) at every measured fraction.}\label{tab:fractions-capacity2048}\\
\toprule
Method / phase & 0\% &10\%&25\%&50\%&75\%&90\%&100\%\\\midrule
\endfirsthead
\toprule Method / phase &0\%&10\%&25\%&50\%&75\%&90\%&100\%\\\midrule\endhead
Joint / re-enc. & 0.03 & 9.08 & 21.44 & 40.46 & 58.71 & 68.84 & 75.58\\
XBT / re-enc. & 63.84 & 53.49 & 50.51 & 55.05 & 62.84 & 68.27 & 72.49\\
MKP / re-enc. & 65.51 & 64.80 & 64.97 & 67.03 & 69.98 & 72.38 & 74.04\\
SLERP / re-enc. & 66.82 & 66.66 & 67.40 & 69.48 & 72.07 & 74.18 & 75.58\\
BT$^{2*}$ / re-enc. & 67.77 & 59.17 & 57.02 & 60.57 & 66.53 & 70.98 & 74.12\\
RACT$^{*}$ / re-enc. & 62.96 & 63.68 & 64.57 & 66.26 & 68.56 & 70.21 & 71.72\\
RankMerge / re-enc. & 61.05 & 61.35 & 61.87 & 63.43 & 66.68 & 70.21 & 75.58\\
MCT / re-enc. & 67.79 & 68.15 & 68.92 & 70.43 & 71.90 & 72.95 & 73.80\\
FastFill$^{*}$ / conv. & 0.03 & 8.73 & 20.11 & 36.87 & 52.51 & 60.91 & 66.52\\
FastFill$^{*}$ / re-enc. & 66.52 & 17.98 & 23.50 & 40.86 & 58.78 & 68.86 & 75.58\\
$\lambda$-Orth / conv. & 66.68 & 66.25 & 66.01 & 66.85 & 68.29 & 69.88 & 71.11\\
$\lambda$-Orth / re-enc. & 71.11 & 71.16 & 71.48 & 72.42 & 73.45 & 74.30 & 74.86\\
BiCT$^{*\dagger}$ / conv. & 42.36 & 26.43 & 28.17 & 39.72 & 52.80 & 60.51 & 65.46\\
BiCT$^{*\dagger}$ / re-enc. & 65.46 & 65.99 & 66.82 & 68.21 & 69.61 & 70.78 & 71.80\\
EC$^{\dagger}$ / conv. & 0.03 & 8.94 & 20.92 & 39.20 & 56.45 & 65.89 & 72.13\\
EC$^{\dagger}$ / re-enc. & 72.13 & 72.45 & 72.90 & 73.59 & 74.51 & 75.09 & 75.58\\
RELAY--P000 / re-enc. & 60.46 & 27.08 & 27.48 & 40.95 & 56.21 & 64.99 & 70.90\\
RELAY--P000 + Calibration / re-enc. & 60.46 & 60.57 & 61.31 & 63.60 & 66.41 & 68.67 & 70.90\\
RELAY--P111 / re-enc. & 69.71 & 51.54 & 50.22 & 55.74 & 64.07 & 69.55 & 73.54\\
RELAY--P111 + Calibration / re-enc. & 69.71 & 67.81 & 66.21 & 64.26 & 65.59 & 68.65 & 73.54\\
RELAY--P100 / re-enc. & 69.71 & 69.58 & 69.67 & 70.17 & 71.86 & 73.17 & 74.77\\
RELAY--P100 + Calibration / re-enc. & 69.71 & 69.53 & 69.77 & 70.48 & 72.23 & 73.51 & 74.77\\
\bottomrule\end{longtable}
\endgroup

Removing the query projection changes the query function at zero steps.
The following TEST values report the score before removal, immediately
after removal, and after training the refined query.
\begin{center}\small\begin{tabular}{llrrr}\toprule Setting & Mask & Drop projection, 0 steps & After refinement & \shortstack{Backfill end\\before removal}\\\midrule
2B-to-2B & P100 & 72.99 & 73.46 & 73.59\\
2B-to-2B & P110 & 72.32 & 73.25 & 72.69\\
CLIP-to-2B & P100 & 67.39 & 72.18 & 69.22\\
CLIP-to-2B & P110 & 66.96 & 71.72 & 70.44\\
2B-to-8B & P100 & 73.40 & 75.57 & 74.77\\
2B-to-8B & P110 & 71.00 & 74.44 & 73.54\\
\bottomrule\end{tabular}\end{center}

The same-base P100 refinement can reduce R@1 even though projection removal and
retuning are implemented correctly. We report that executed endpoint,
rather than retrospectively choosing the better of the backfill and refined endpoints.

\subsubsection{Per-dataset and secondary retrieval metrics}
AUC columns below use R@1. Each Backfill end/Final system cell reports R@1/R@5/R@10/mAP@5,
in percent, for all four masks and both DEV/TEST pools. Calibration leaves
these homogeneous endpoints unchanged. Neither a
macro average nor a good final point replaces the full migration curve.
\begingroup\small
\begin{longtable}{llllrrr}
\caption{Current projection masks by dataset and split (\%). Each endpoint reports all four retrieval metrics; AUC uses R@1.}\label{tab:projection-pools}\\
\toprule Setting & Mask & Pool & Split & Raw / calibrated AUC & \shortstack{Backfill end\\(R1/R5/R10/AP5)} & \shortstack{Final system\\(R1/R5/R10/AP5)}\\\midrule
\endfirsthead
\toprule Setting & Mask & Pool & Split & Raw / calibrated AUC & Backfill end & Final system\\\midrule\endhead
2B-to-2B & P000 & COCO & DEV & 58.63 / 59.54 & 61.81 / 84.25 / 90.52 / 70.53 & 61.73 / 84.41 / 90.56 / 70.54\\
2B-to-2B & P000 & Flickr & DEV & 84.69 / 85.81 & 87.18 / 97.36 / 98.90 / 91.47 & 86.98 / 97.20 / 98.88 / 91.34\\
2B-to-2B & P000 & COCO & TEST & 58.07 / 58.91 & 61.15 / 83.89 / 90.40 / 70.08 & 61.49 / 84.10 / 90.54 / 70.30\\
2B-to-2B & P000 & Flickr & TEST & 84.15 / 85.10 & 86.10 / 97.42 / 98.84 / 90.86 & 86.22 / 97.44 / 98.90 / 90.92\\
2B-to-2B & P111 & COCO & DEV & 50.77 / 57.51 & 60.83 / 83.82 / 90.22 / 69.76 & 60.13 / 83.42 / 90.20 / 69.18\\
2B-to-2B & P111 & Flickr & DEV & 78.63 / 84.27 & 86.09 / 97.36 / 98.82 / 90.79 & 85.62 / 97.28 / 98.78 / 90.47\\
2B-to-2B & P111 & COCO & TEST & 50.30 / 57.27 & 60.30 / 83.49 / 90.03 / 69.26 & 59.58 / 83.27 / 89.91 / 68.78\\
2B-to-2B & P111 & Flickr & TEST & 78.12 / 83.33 & 85.08 / 97.34 / 98.58 / 90.18 & 84.50 / 97.06 / 98.74 / 89.77\\
2B-to-2B & P110 & COCO & DEV & 50.77 / 57.51 & 60.83 / 83.82 / 90.22 / 69.76 & 61.26 / 84.16 / 90.36 / 70.14\\
2B-to-2B & P110 & Flickr & DEV & 78.63 / 84.27 & 86.09 / 97.36 / 98.82 / 90.79 & 86.19 / 97.30 / 98.95 / 90.88\\
2B-to-2B & P110 & COCO & TEST & 50.30 / 57.27 & 60.30 / 83.49 / 90.03 / 69.26 & 60.81 / 83.81 / 90.25 / 69.79\\
2B-to-2B & P110 & Flickr & TEST & 78.12 / 83.33 & 85.08 / 97.34 / 98.58 / 90.18 & 85.70 / 97.28 / 98.76 / 90.55\\
2B-to-2B & P100 & COCO & DEV & 59.16 / 59.68 & 61.44 / 84.10 / 90.53 / 70.31 & 61.36 / 84.17 / 90.22 / 70.18\\
2B-to-2B & P100 & Flickr & DEV & 85.28 / 85.74 & 86.67 / 97.46 / 98.80 / 91.21 & 86.84 / 97.26 / 98.90 / 91.25\\
2B-to-2B & P100 & COCO & TEST & 58.87 / 59.35 & 61.37 / 83.82 / 90.23 / 70.08 & 61.17 / 83.83 / 90.21 / 69.97\\
2B-to-2B & P100 & Flickr & TEST & 84.61 / 85.12 & 85.82 / 97.40 / 98.80 / 90.67 & 85.76 / 97.34 / 99.00 / 90.59\\
CLIP-to-2B & P000 & COCO & DEV & 32.62 / 44.39 & 54.35 / 80.62 / 88.31 / 64.47 & 58.49 / 82.79 / 89.68 / 67.91\\
CLIP-to-2B & P000 & Flickr & DEV & 48.65 / 73.27 & 80.79 / 95.88 / 97.99 / 87.03 & 83.85 / 96.65 / 98.52 / 89.17\\
CLIP-to-2B & P000 & COCO & TEST & 32.34 / 44.06 & 53.89 / 80.20 / 88.13 / 64.06 & 58.05 / 82.47 / 89.13 / 67.53\\
CLIP-to-2B & P000 & Flickr & TEST & 47.97 / 72.22 & 79.80 / 95.88 / 98.20 / 86.44 & 83.22 / 96.60 / 98.36 / 88.80\\
CLIP-to-2B & P111 & COCO & DEV & 35.22 / 47.67 & 58.64 / 82.84 / 89.65 / 67.98 & 59.17 / 83.44 / 89.94 / 68.56\\
CLIP-to-2B & P111 & Flickr & DEV & 52.62 / 77.29 & 84.04 / 96.80 / 98.50 / 89.40 & 84.58 / 96.92 / 98.54 / 89.79\\
CLIP-to-2B & P111 & COCO & TEST & 34.71 / 47.04 & 57.64 / 82.41 / 89.29 / 67.29 & 58.35 / 82.94 / 89.67 / 67.89\\
CLIP-to-2B & P111 & Flickr & TEST & 52.08 / 75.91 & 83.24 / 96.58 / 98.60 / 88.82 & 83.98 / 96.78 / 98.72 / 89.35\\
CLIP-to-2B & P110 & COCO & DEV & 35.22 / 47.67 & 58.64 / 82.84 / 89.65 / 67.98 & 60.21 / 83.70 / 90.13 / 69.30\\
CLIP-to-2B & P110 & Flickr & DEV & 52.62 / 77.29 & 84.04 / 96.80 / 98.50 / 89.40 & 85.23 / 97.14 / 98.66 / 90.23\\
CLIP-to-2B & P110 & COCO & TEST & 34.71 / 47.04 & 57.64 / 82.41 / 89.29 / 67.29 & 59.26 / 83.35 / 89.79 / 68.70\\
CLIP-to-2B & P110 & Flickr & TEST & 52.08 / 75.91 & 83.24 / 96.58 / 98.60 / 88.82 & 84.18 / 97.04 / 98.64 / 89.57\\
CLIP-to-2B & P100 & COCO & DEV & 38.35 / 49.15 & 57.57 / 82.15 / 89.28 / 67.10 & 59.98 / 83.53 / 90.22 / 69.07\\
CLIP-to-2B & P100 & Flickr & DEV & 61.65 / 78.09 & 82.64 / 96.61 / 98.44 / 88.47 & 85.46 / 97.00 / 98.66 / 90.26\\
CLIP-to-2B & P100 & COCO & TEST & 37.66 / 48.21 & 56.54 / 81.77 / 88.81 / 66.32 & 59.61 / 83.30 / 89.74 / 68.83\\
CLIP-to-2B & P100 & Flickr & TEST & 60.77 / 76.32 & 81.90 / 96.46 / 98.44 / 87.89 & 84.74 / 97.18 / 98.62 / 89.92\\
2B-to-8B & P000 & COCO & DEV & 36.16 / 51.15 & 59.17 / 83.22 / 89.85 / 68.50 & 62.70 / 85.11 / 91.15 / 71.35\\
2B-to-8B & P000 & Flickr & DEV & 55.45 / 79.00 & 84.67 / 96.90 / 98.42 / 89.81 & 86.88 / 97.53 / 98.90 / 91.36\\
2B-to-8B & P000 & COCO & TEST & 35.55 / 50.55 & 58.19 / 82.41 / 89.68 / 67.62 & 61.85 / 84.51 / 90.93 / 70.70\\
2B-to-8B & P000 & Flickr & TEST & 54.56 / 77.79 & 83.62 / 96.96 / 98.70 / 89.24 & 86.56 / 97.84 / 98.96 / 91.29\\
2B-to-8B & P111 & COCO & DEV & 45.79 / 53.36 & 61.58 / 84.50 / 90.75 / 70.51 & 60.78 / 84.40 / 90.69 / 69.95\\
2B-to-8B & P111 & Flickr & DEV & 73.52 / 81.85 & 86.27 / 97.75 / 98.95 / 91.10 & 85.70 / 97.34 / 98.72 / 90.58\\
2B-to-8B & P111 & COCO & TEST & 45.38 / 52.72 & 60.83 / 84.00 / 90.63 / 69.90 & 60.38 / 83.65 / 90.42 / 69.46\\
2B-to-8B & P111 & Flickr & TEST & 72.81 / 80.57 & 86.24 / 97.46 / 98.94 / 90.96 & 85.78 / 97.26 / 98.84 / 90.58\\
2B-to-8B & P110 & COCO & DEV & 45.79 / 53.36 & 61.58 / 84.50 / 90.75 / 70.51 & 62.51 / 84.87 / 91.09 / 71.31\\
2B-to-8B & P110 & Flickr & DEV & 73.52 / 81.85 & 86.27 / 97.75 / 98.95 / 91.10 & 86.80 / 97.87 / 98.86 / 91.44\\
2B-to-8B & P110 & COCO & TEST & 45.38 / 52.72 & 60.83 / 84.00 / 90.63 / 69.90 & 61.98 / 84.88 / 90.97 / 70.89\\
2B-to-8B & P110 & Flickr & TEST & 72.81 / 80.57 & 86.24 / 97.46 / 98.94 / 90.96 & 86.90 / 97.66 / 98.80 / 91.43\\
2B-to-8B & P100 & COCO & DEV & 58.14 / 58.50 & 62.86 / 85.27 / 91.22 / 71.63 & 64.08 / 85.70 / 91.57 / 72.48\\
2B-to-8B & P100 & Flickr & DEV & 84.88 / 85.20 & 87.36 / 97.55 / 98.84 / 91.60 & 87.99 / 98.01 / 99.07 / 92.17\\
2B-to-8B & P100 & COCO & TEST & 57.80 / 58.00 & 62.55 / 84.74 / 90.94 / 71.20 & 63.30 / 85.37 / 91.32 / 71.92\\
2B-to-8B & P100 & Flickr & TEST & 84.04 / 84.25 & 86.98 / 97.74 / 99.00 / 91.54 & 87.84 / 97.72 / 99.10 / 92.03\\
\bottomrule\end{longtable}\endgroup

\subsubsection{Completed matched controls}
All main-table seed 42 method slots are complete. The two Qwen settings use
matching r32 Joint teachers for MKP, full XBT, and FastFill; CLIP RACT reuses
its accepted same-instance retrieval adaptation. Native scoring is retained
throughout. MKP and XBT end at their transformed compatible pairs, whereas
FastFill's re-encoding endpoint is its native teacher.
Table~\ref{tab:completed-baseline-pools}
provides the per-dataset results for these eight completed controls.
\Cref{tab:fractions-same2048,tab:fractions-clip768,tab:fractions-capacity2048}
report native seven-point curves for every method in
\cref{tab:projection-main}, with separate rows for conversion and
re-encoding. Supplementary FastFill uncertainty-order rows retain the
same pure endpoints. Table~\ref{tab:projection-main} uses the common order.
\begingroup\small\setlength{\tabcolsep}{3pt}
\begin{longtable}{lllrrl}
\caption{Newly completed matched controls, seed 42 TEST (\%). Phase start and Phase end refer to the listed phase; Phase end lists R@1/R@5/R@10/mAP@5. Conversion end equals re-encoding start. The uncertainty-order FastFill rows are supplementary; the main table uses the common hash order. The re-encoding endpoint is Backfill end, with no query refinement for these controls.}\label{tab:completed-baseline-pools}\\
\toprule Setting & Method / phase & Dataset & \shortstack{Phase\\start} & \shortstack{Phase\\AUC} & \shortstack{Phase end\\(R1/R5/R10/AP5)}\\\midrule\endfirsthead
\toprule Setting & Method / phase & Dataset & \shortstack{Phase\\start} & \shortstack{Phase\\AUC} & \shortstack{Phase end\\(R1/R5/R10/AP5)}\\\midrule\endhead
2B-to-2B & XBT & COCO & 49.58 & 44.24 & 58.69 / 82.96 / 89.90 / 68.10\\
2B-to-2B & XBT & Flickr & 77.94 & 71.24 & 84.06 / 96.80 / 98.50 / 89.45\\
2B-to-2B & MKP-Adapter & COCO & 51.79 & 54.39 & 60.65 / 83.49 / 89.87 / 69.55\\
2B-to-2B & MKP-Adapter & Flickr & 78.34 & 80.72 & 85.76 / 97.28 / 98.76 / 90.61\\
2B-to-2B & FastFill$^{*}$ / conv. & COCO & 54.18 & 33.36 & 53.63 / 80.22 / 87.77 / 63.81\\
2B-to-2B & FastFill$^{*}$ / conv. & Flickr & 80.72 & 51.93 & 81.14 / 96.06 / 98.08 / 87.34\\
2B-to-2B & FastFill$^{*}$ / re-enc. & COCO & 53.63 & 36.27 & 61.71 / 84.73 / 90.79 / 70.69\\
2B-to-2B & FastFill$^{*}$ / re-enc. & Flickr & 81.14 & 50.92 & 86.34 / 97.46 / 98.74 / 91.06\\
2B-to-2B & FastFill$^{*}$ / uncertainty & COCO & 53.63 & 33.66 & 61.71 / 84.73 / 90.79 / 70.69\\
2B-to-2B & FastFill$^{*}$ / uncertainty & Flickr & 81.14 & 50.47 & 86.34 / 97.46 / 98.74 / 91.06\\
\midrule
CLIP-to-2B & RACT$^{*}$ & COCO & 36.81 & 35.60 & 56.64 / 81.92 / 88.82 / 66.43\\
CLIP-to-2B & RACT$^{*}$ & Flickr & 65.06 & 56.66 & 83.06 / 96.48 / 98.52 / 88.75\\
\midrule
2B-to-8B & Joint & COCO & 0.03 & 34.17 & 63.36 / 85.66 / 91.62 / 72.12\\
2B-to-8B & Joint & Flickr & 0.04 & 45.16 & 87.80 / 97.70 / 99.06 / 92.01\\
2B-to-8B & XBT & COCO & 49.40 & 44.42 & 59.44 / 83.74 / 90.26 / 68.92\\
2B-to-8B & XBT & Flickr & 78.28 & 72.52 & 85.54 / 97.14 / 98.64 / 90.33\\
2B-to-8B & MKP-Adapter & COCO & 52.17 & 54.42 & 61.29 / 84.25 / 90.40 / 70.27\\
2B-to-8B & MKP-Adapter & Flickr & 78.84 & 81.33 & 86.78 / 97.38 / 98.78 / 91.21\\
2B-to-8B & FastFill$^{*}$ / conv. & COCO & 0.03 & 29.94 & 53.53 / 80.26 / 87.80 / 63.88\\
2B-to-8B & FastFill$^{*}$ / conv. & Flickr & 0.04 & 41.61 & 79.52 / 95.90 / 98.12 / 86.31\\
2B-to-8B & FastFill$^{*}$ / re-enc. & COCO & 53.53 & 37.20 & 63.36 / 85.66 / 91.62 / 72.12\\
2B-to-8B & FastFill$^{*}$ / re-enc. & Flickr & 79.52 & 52.06 & 87.80 / 97.70 / 99.06 / 92.01\\
2B-to-8B & FastFill$^{*}$ / uncertainty & COCO & 53.53 & 35.12 & 63.36 / 85.66 / 91.62 / 72.12\\
2B-to-8B & FastFill$^{*}$ / uncertainty & Flickr & 79.52 & 51.95 & 87.80 / 97.70 / 99.06 / 92.01\\
\bottomrule\end{longtable}\endgroup

\Needspace{17\baselineskip}
\subsubsection{Paired training repeats}
\label{sec:projection-seed-repeats}
The main table uses training seed 42 and replacement seed 20261104.
We repeat CLIP-to-Qwen 2B and Qwen 2B-to-8B with seeds 43 and 44.
For each seed, P100 reuses its query-adaptation checkpoint and trains
document adaptation and query refinement for 960 steps each. MCT uses the
corresponding same-seed Joint2880 teacher. We fit separate isotonic maps
for each method and seed on DEV, then use the same seven-fraction
replacement order on TEST. Query refinement is evaluated after full
backfill and is excluded from AUC.

For MCT, the same DEV selection rule chooses epochs 50, 25, and 25 for
Qwen 2B-to-8B seeds 42, 43, and 44. DEV selects epoch 50 for both
CLIP-to-Qwen repeats. \Cref{tab:projection-seed-repeats} reports their
absolute scores.
\begin{center}\small\setlength{\tabcolsep}{4pt}
\begin{tabular}{ll l r r r r r}\toprule
Upgrade & Seed & Method & Entry & \shortstack{AUC\\Native / cal.} & \shortstack{Minimum\\Native / cal.} & \shortstack{Backfill\\end} & \shortstack{After\\refinement}\\\midrule
CLIP-to-2B & 43 & RELAY--P100 & 59.31 & 49.31 / 62.39 & 37.14 / 58.92 & 69.22 & 71.98\\
CLIP-to-2B & 43 & MCT & 50.99 & 59.51 / 60.22 & 50.99 / 50.99 & 71.49 & --\\
CLIP-to-2B & 44 & RELAY--P100 & 59.38 & 48.66 / 62.18 & 35.51 / 58.84 & 69.35 & 72.52\\
CLIP-to-2B & 44 & MCT & 50.71 & 59.39 / 60.15 & 50.71 / 50.71 & 72.03 & --\\
\midrule
2B-to-8B & 43 & RELAY--P100 & 69.41 & 70.37 / 70.91 & 69.11 / 69.25 & 74.71 & 75.54\\
2B-to-8B & 43 & MCT & 67.64 & 70.31 / 70.42 & 67.64 / 67.64 & 73.58 & --\\
2B-to-8B & 44 & RELAY--P100 & 69.90 & 70.42 / 71.07 & 69.44 / 69.70 & 74.42 & 75.41\\
2B-to-8B & 44 & MCT & 68.01 & 70.73 / 70.88 & 68.01 / 68.01 & 74.10 & --\\
\bottomrule\end{tabular}
\captionof{table}{\textbf{Additional seeds for direct re-encoding.} Equal-dataset TEST macro R@1 (\%) on COCO and Flickr30K. AUC integrates the seven backfill fractions, and Minimum is the lowest measured R@1. Entry and Backfill end are the homogeneous-gallery endpoints. After refinement reports the separate P100 query endpoint.}
\label{tab:projection-seed-repeats}\end{center}

For Qwen 2B-to-2B, RELAY and RACT share the initial LoRA weights within
each additional seed. Seeds 43 and 44 reuse seed 42's training data and
sampling order, and each model has its own DEV-fitted calibration maps.
RELAY leads RACT in native and calibrated AUC under each of the six
replacement orders for both additional seeds. On the main replacement
order, RELAY's entry R@1 leads by 4.01 and 3.89 points for seeds 43 and 44.
At full backfill, R@1 with the same deployed query trails RACT by 0.57
and 0.80 points, respectively.
\begin{center}\small
\begin{tabular}{llrrrr}\toprule
Upgrade and comparator & Scoring & Seed 42 & Seed 43 & Seed 44 & Mean $\pm$ SD \\
\midrule
2B-to-2B vs. RACT & Native & $+1.45$ & $+1.44$ & $+1.29$ & $+1.40\pm0.09$ \\
 & Calibrated & $+1.90$ & $+1.87$ & $+1.88$ & $+1.88\pm0.02$ \\
CLIP-to-2B vs. MCT & Native & $-10.21$ & $-10.20$ & $-10.73$ & $-10.38\pm0.31$ \\
 & Calibrated & $+2.05$ & $+2.17$ & $+2.03$ & $+2.08\pm0.08$ \\
2B-to-8B vs. MCT & Native & $+0.43$ & $+0.06$ & $-0.31$ & $+0.06\pm0.37$ \\
 & Calibrated & $+0.56$ & $+0.49$ & $+0.20$ & $+0.42\pm0.20$ \\
\bottomrule\end{tabular}
\captionof{table}{\textbf{Paired TEST re-encoding AUC differences across training seeds.}
Entries are RELAY minus the comparator, in percentage points, on the
main replacement order. SD is the sample standard deviation across the
three paired differences.}
\label{tab:paired-training-repeats}
\end{center}

\Needspace{8\baselineskip}
\subsubsection{Fixed-query document training and post-hoc mapping}
\label{sec:fixed-query-control}
Both arms serve the exact P100 $Q_1$ and start from the same $D_0$.
At each fraction, the same image IDs are replaced by either P100 $D_1$
vectors or mapped Joint document vectors. We fit the map on 122,856
distinct training images, pairing each frozen Joint document vector
with its $D_0$ vector. For the three ridge candidates, the diagonal
penalty is $\alpha\,\operatorname{tr}(X^\top X)/d$ with
$\alpha\in\{0.001,0.01,0.1\}$, where $X$ stacks Joint document vectors.
We also fit an orthogonal Procrustes map. We normalize mapped vectors
before scoring. The mapping fit uses document pairs alone.

We select a map by native macro AUC on the DEV queries held out from
isotonic fitting, breaking ties by the sampled minimum, the 100\%
endpoint, and then the fixed candidate order (ridge penalties in
ascending order, then Procrustes). We fit score maps on the other DEV
queries with the protocol in \cref{sec:score-calibration}. The two arms
share the old-version map and each has its own new-version map.
\Cref{tab:fixed-query-absolute,fig:fixed-query-curves} report TEST
results with $Q_1$ through the 100\% endpoint.

\begin{center}\small
\begin{tabular}{llrrrrrr}\toprule
Upgrade & Mapped Joint fit & Native AUC & Calib. AUC & Native min. & Calib. min. & 0\% R@1 & 100\% R@1 \\
\midrule
2B-to-2B & Ridge (0.001) & 71.74 / 71.24 & 72.23 / 71.83 & 71.09 / 70.56 & 71.28 / 71.28 & 71.28 & 73.59 / 72.66 \\
CLIP-to-2B & Ridge (0.01) & 49.22 / 60.72 & 62.27 / 61.26 & 36.60 / 59.24 & 58.99 / 59.74 & 59.74 & 69.22 / 64.16 \\
2B-to-8B & Procrustes & 70.92 / 69.87 & 71.13 / 70.16 & 69.58 / 69.11 & 69.53 / 69.70 & 69.71 & 74.77 / 71.70 \\
\bottomrule\end{tabular}
\captionof{table}{\textbf{Fixed-query results on TEST (\%).}
Paired cells show RELAY / mapped Joint. At 0\%, both arms share the
same value. At 100\%, both still use $Q_1$. Ridge values are $\alpha$.}
\label{tab:fixed-query-absolute}
\end{center}

\begin{center}
\includegraphics[width=\linewidth]{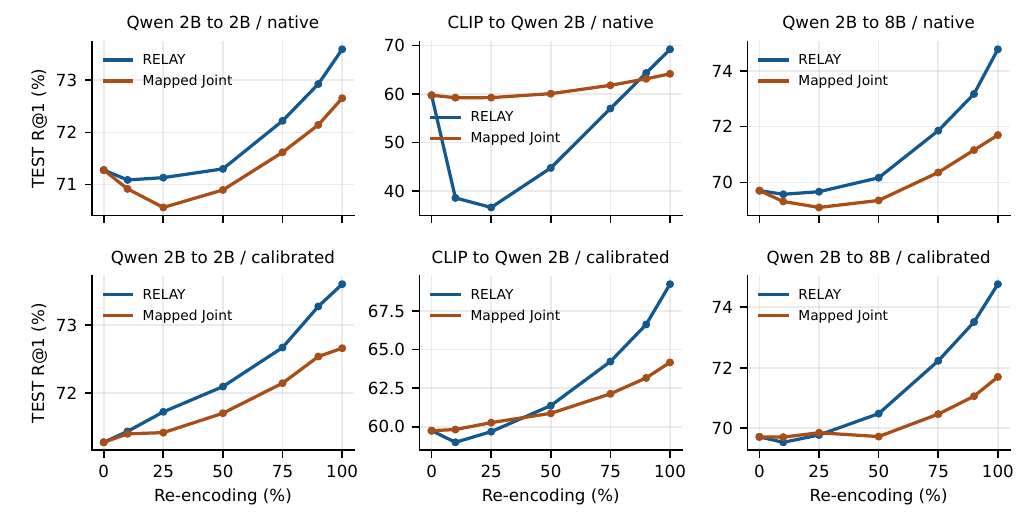}
\captionof{figure}{\textbf{Fixed-query comparison on TEST.}
Each panel compares P100 $D_1$ with mapped Joint document vectors under
the original replacement order. Both arms share $Q_1$ and the old
gallery, so they coincide at 0\%. Markers show the seven AUC fractions.}
\label{fig:fixed-query-curves}
\end{center}

\Needspace{8\baselineskip}
\subsubsection{Sensitivity to gallery replacement order}
\label{sec:replacement-order}
We reuse the trained encoders, cached features, and DEV-fitted maps
from \cref{tab:projection-main}. Within each dataset and order, all
compared methods replace the same nested sequence of document IDs.
We rank IDs by SHA-256 of \texttt{seed:ID}, using the original seed
20261104 and five additional seeds (101--105). Hash ties use the original
document row.
\Cref{tab:replacement-order-paired,fig:replacement-order-curves} give
paired AUC differences and all six curves for the selected comparators.
The same seven fractions enter every AUC. The additional CLIP-to-Qwen
samples in \cref{fig:clip-early-backfill} measure the early minimum.

\begin{center}\small
\begin{tabular}{llrrrrrr}\toprule
Upgrade & Comparator and scoring & 20261104 & 101 & 102 & 103 & 104 & 105 \\
\midrule
2B-to-2B & RACT, native & +1.45 & +1.55 & +1.52 & +1.57 & +1.42 & +1.38 \\
2B-to-2B & RACT, calibrated & +1.90 & +1.94 & +1.78 & +1.95 & +1.85 & +1.78 \\
CLIP-to-2B & MCT, native & -10.21 & -10.76 & -10.47 & -10.50 & -10.20 & -10.54 \\
CLIP-to-2B & BT$^2$, calibrated & -2.99 & -2.84 & -2.74 & -3.03 & -2.67 & -3.13 \\
2B-to-8B & MCT, native & +0.43 & +0.41 & +0.44 & +0.25 & +0.41 & +0.31 \\
2B-to-8B & MCT, calibrated & +0.56 & +0.58 & +0.63 & +0.45 & +0.71 & +0.52 \\
\bottomrule\end{tabular}
\captionof{table}{\textbf{Paired TEST AUC differences by replacement seed.}
Each cell is RELAY minus the named comparator in percentage points.
The encoders use training seed 42 throughout.}
\label{tab:replacement-order-paired}
\end{center}

\begin{center}
\includegraphics[width=\linewidth]{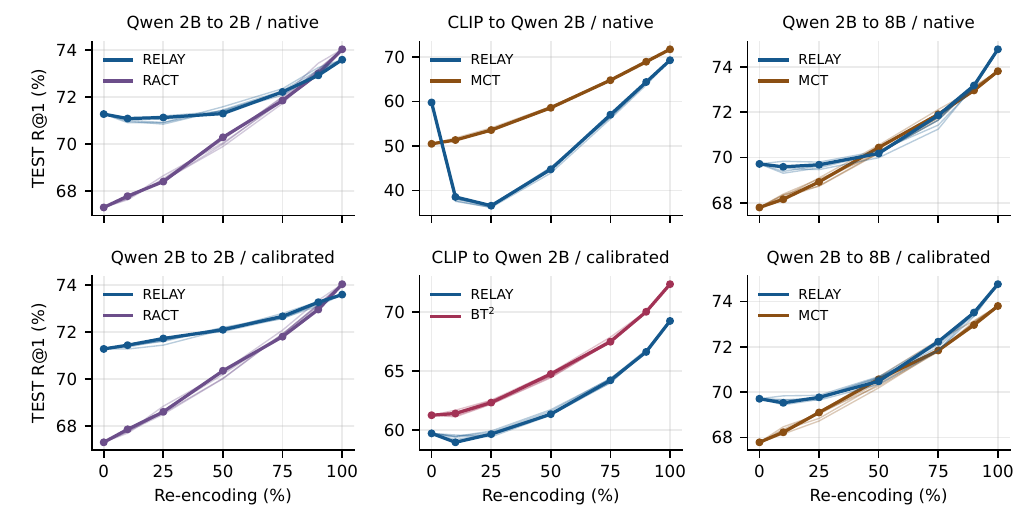}
\captionof{figure}{\textbf{Direct re-encoding under six replacement orders.}
Each panel shows six curves per method. Markers identify the original
order (20261104). Thin lines show the five additional orders.}
\label{fig:replacement-order-curves}
\end{center}

\begin{center}
\includegraphics[width=0.90\linewidth]{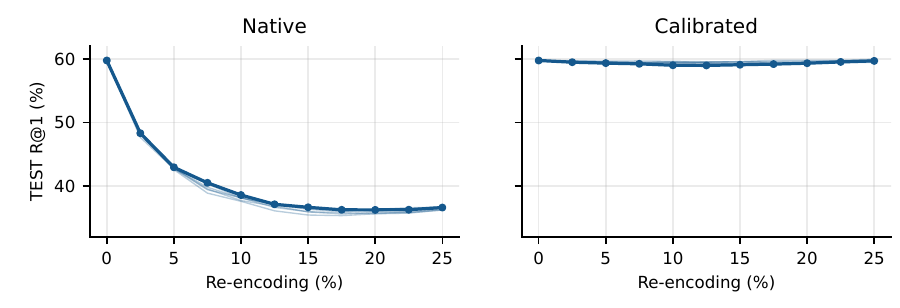}
\captionof{figure}{\textbf{Early CLIP-to-Qwen backfill.}
RELAY uses fixed features across the six orders. We evaluate every
2.5 percentage points through 25\% backfill. Markers identify the
original order. Thin lines show the five additional orders.}
\label{fig:clip-early-backfill}
\end{center}

\begin{center}\small
\begin{tabular}{lrr}\toprule
Upgrade & COCO & Flickr30K \\\midrule
2B-to-2B & $58.87\to59.35$ & $84.61\to85.12$ \\
CLIP-to-2B & $37.66\to48.21$ & $60.77\to76.32$ \\
2B-to-8B & $57.80\to58.00$ & $84.04\to84.25$ \\
\bottomrule\end{tabular}
\captionof{table}{RELAY (P100) backfill AUC on each test set (\%). Arrows
pair native and calibrated AUC on the same features.}
\label{tab:calibration-pool-summary}
\end{center}

\subsubsection{Training and feature-encoding cost boundaries}
Main Table~\ref{tab:projection-cost} includes all adopted training stages,
not just the new P100 branches. It counts each actual successful model
process once, including internal load/save, and counts reused query adaptation once per
independent recipe. Model steps are 2880/960/960 (614,400 image-group draw
occurrences in total), versus Joint2880's 368,640. Both have an effective
batch of 128 images. Equal nominal GradCache chunk 16 does not imply equal
execution: Joint replays both independently trainable role LoRAs; RELAY
uses its specified cached or online no-gradient opposite role. Schedules
also differ. The percentages are therefore observed recipe costs, not a
causal estimate of one-sided backpropagation alone.

Feature encoding is accounted for separately from training. CLIP P100
requires new query and document TRAIN features. Qwen stages 2 and 3
evaluate the frozen role online. The creation cost of the shared
old-document TRAIN cache is unknown. Joint's extra TRAIN features support
post-hoc mapping baselines. The available encoding measurements cover
different data splits and process boundaries across recipes, so the
reported savings apply to training. Production backfill, index maintenance,
and online serving costs are outside these totals.

\Needspace{6\baselineskip}
\subsection{Data and training examples}
\label{sec:data-splits}
COCO training and DEV derive from the 2017 training annotations; TEST uses all
5,000 val 2017 images. Candidate DEV images are ordered by
SHA-256(20261104:coco:image-ID), after excluding source-photo and exact-byte
overlaps with Flickr held-out images and final TEST. The first 5,000 form DEV.
This is a COCO-2017-based partition with a custom DEV holdout.
Flickr splits come from the original dataset\_flickr 30k.json Karpathy file~\cite{karpathy2015alignments};
we use its 1,014-image validation split as the development set and its 1,000-image
test split as the test set. All original
evaluation captions are without text deduplication. COCO therefore
has 25,018/25,014 DEV/TEST caption rows rather than exactly five per image.
Each caption row has its own query ID and one image-level relevance target;
identical caption strings do not merge query identities or targets.

Training excludes every image with a held-out source-photo identity or exact
original-image-byte SHA-256 match, including overlaps across COCO/Flickr.
Source-photo identity is COCO's original Flickr photo ID versus Flickr's
filename stem; this is not perceptual near-duplicate filtering. The resulting
training pools contain 113,280 COCO and 28,990 Flickr images. Only the first
five COCO training annotations, sorted by annotation ID, enter training.
The 960-step 4:1 schedule draws 98,304/24,576 logical COCO/Flickr image groups,
respectively; it is a fixed sampled pass rather than one epoch over their
entire union. Each global step draws one dataset, with no duplicate image
bytes or normalized caption shared between different images in that batch.
The normalizer for this sampling check is Unicode NFKC, casefold, and whitespace
collapse; it does not rewrite the actual input text. Within-image duplicate
captions can remain and share the correct target.

These checks remove the specified overlaps among our training and evaluation
splits; public images may still have appeared in backbone pretraining.

\Needspace{6\baselineskip}
\subsection{Shared implementation and model selection}
In this section, ``Qwen'' denotes the 2B-to-2B upgrade. The 8B recipes
are specified in the preceding three-scenario section.

\Needspace{4\baselineskip}
\subsubsection{Encoder training}
The public pipeline uses separate query and document LoRA adapters over a frozen
Qwen3-VL-Embedding backbone: 2B in the earlier controls, and 2B or 8B in the current study.
The target regex is restricted to language-model modules named
\texttt{q\_proj},\texttt{k\_proj},\texttt{v\_proj},\texttt{o\_proj}, or
\texttt{gate}; the vision encoder lies outside this pattern. Rank, alpha, and dropout are 32,64,0.05.
Training uses BF16, AdamW with betas(0.9,0.95), epsilon $10^{-8}$, weight decay 0.1,
gradient clipping 1, warmup ratio 0.001, and cosine schedules. Per-device image
group batch is 16, world size 8, and gradient accumulation 1. GradCache replay
chunks contain 16 expanded rows. Each group has five captions and one image;
the effective batch therefore has 128 image groups and 640 text queries, not
768 independent positive image classes. One 122,880-group pass has 960
optimizer steps. The group sequence and sampling manifest are fixed and repeat
without reshuffling. The prompt is ``Represent the user's input.''

\paragraph{Image processing.}
All trainable Qwen roles share the following Qwen image pipeline; the frozen
CLIP source retains its own preprocessing described below. Training loads original
images through the Swift Qwen3-VL template with a 262,144-pixel cap
and maximum sequence length 4096; the registered image processor uses
channel mean and standard deviation $(0.5,0.5,0.5)$ after pixel rescaling.
For cached Qwen features and evaluation, the SDK first decodes the original
bytes, resizes without cropping to dimensions divisible by 32 with area
between 4,096 and 262,144 pixels, and encodes JPEG at quality 95 before
the Qwen processor. Resizing uses OpenCV area interpolation when shrinking
and cubic interpolation when enlarging, with Pillow as a decode fallback.
This SDK JPEG step is an inference/cache boundary, not an added training
augmentation.

The frozen CLIP model is \texttt{openai/clip-vit-large-patch14}, revision
\texttt{32bd64288804d66eefd0ccbe215aa642df71cc41}. Its slow
\texttt{CLIPProcessor} consumes original RGB images directly, with no Qwen
resize or JPEG re-encoding: bicubic shorter-edge resize to 224, followed by
a $224\times224$ center crop. After $1/255$ rescaling, the channel means are
$(0.48145466,0.4578275,0.40821073)$ and standard deviations
$(0.26862954,0.26130258,0.27577711)$. These distinct CLIP and Qwen
pipelines are retained for their respective frozen and trainable models.

The two Qwen-to-Qwen settings use equally weighted 2048/1024/768/512 prefix objectives; evaluation uses
the normalized 2048-vector. CLIP-to-Qwen trains and evaluates the normalized 768
prefix. There are no weak-positive or query--query terms on these paired
caption examples. Stages 1--3 update the selected role only; the other role is
consumed through its frozen features. Qwen stage 1 uses checkpoint 2880
from a 5760-step cosine trajectory.

\paragraph{Joint-training controls.}
Joint is the conventional simultaneous-update counterpart to the staged
training strategy: the backbone initialization, rank, retrieval objective,
training examples and global batch are shared. We test single learning-rate
schedules at 2880 and 4800 steps and a staged schedule at 4800 steps using
RELAY's phase lengths and learning rates.

\begin{table}[ht]\centering\small
\begin{tabular}{llrrrr}
\toprule Setting & Joint recipe & Base LR(s) & Cosine horizon(s) & DEV & TEST\\\midrule
Qwen & 2880 steps, single schedule & $5\times10^{-4}$ & 2880 & \textbf{74.54} & 74.02\\
Qwen & 4800 steps, single schedule & $5\times10^{-4}$ & 4800 & 73.93 & 73.54\\
Qwen & 4800 steps, staged schedule & $(1,5,5)\times10^{-4}$ & 5760/960/960 & 74.23 & 73.85\\
\midrule
CLIP & 2880 steps, single schedule & $5\times10^{-4}$ & 2880 & \textbf{73.71} & 73.15\\
CLIP & 4800 steps, single schedule & $5\times10^{-4}$ & 4800 & 73.24 & 72.36\\
CLIP & 4800 steps, staged schedule & $(3,5,5)\times10^{-4}$ & 2880/960/960 & 73.14 & 72.51\\
\bottomrule\end{tabular}
\caption{Rank-32 Joint controls, seed 42. DEV and TEST give native
text-to-image R@1 (\%), averaged equally over both datasets. The 2880-step
run gives the highest development-set R@1. All runs start from the pretrained
Qwen backbone with fresh role adapters and update both roles at every step.
The staged schedule carries model weights forward and resets the optimizer
and learning-rate scheduler after steps 2880 and 3840. Qwen's first phase uses
the first 2880 steps of its 5760-step cosine schedule. The remaining
optimizer and batch settings are shared with encoder training above.}
\label{tab:joint-selection}\end{table}

\Needspace{4\baselineskip}
\subsubsection{Development selection}
\label{sec:development-selection}
Candidate ranking and score-map fitting use the development set. The
diagnostic split used to check calibration is part of DEV and remains
available for encoder selection.

Qwen's rank and stage 1 learning-rate/horizon were explored on DEV before the earlier LoRA
representative was fixed.
CLIP stage 1 was selected among 12 checkpoint/horizon/learning-rate
candidates, some sharing a parent trajectory rather than constituting 12 new
full runs. The chosen stage 1 is 2880 steps at $3\times 10^{-4}$.

The complete eligible Qwen set consists of ranks $\{8,16,32,64\}$ at
stage 1 LR $5\times10^{-4}$ and ranks $\{16,32\}$ at LR $10^{-4}$, each with stage 1
prefixes 960/2880 from a 5760-step trajectory and its own 960-step stage 2: 12
eligible chains. Rank 64 with stage 1 prefix 5760 was an extra DEV-only diagnostic,
ineligible for selection. Selection maximizes equal-dataset seven-point DEV
raw R@1 AUC, then floor, Doc100, smaller rank, shorter prefix, lower stage 1 LR,
and lexical case ID. It retains the winning rank/LR's 960/2880 prefix pair;
the winning 2880 chain has DEV AUC 71.5126\%, rank 32, LR $10^{-4}$.

CLIP keeps rank 32 and selects on DEV original-gallery R@1, then earlier actual
step and lower base LR; an unresolved tie is rejected. Its 12 candidates are:
LR $10^{-4}$/horizon 960 at step 960; LR $10^{-4}$/horizon 2880 at
960/1920/2880; LR $10^{-4}$/horizon 5760 at 960/1920/2880/4320/5760; and
LR $3\times10^{-4}$/horizon 2880 at 960/1920/2880. The last trajectory's
2880 checkpoint wins at 55.1677\% DEV entry. These are checkpoints,
not 12 independent full training runs. The two settings thus use different
selection criteria (Qwen downstream migration AUC; CLIP old-library entry),
so the two settings differ in selection objective as well as source model.

For SLERP, we evaluate $\alpha\in\{0,0.1,\ldots,1\}$ separately for each
upgrade. We select the value with the highest DEV Entry R@1 and keep it
fixed throughout TEST migration. Both Qwen-to-Qwen upgrades select
$\alpha=1$, and CLIP-to-Qwen selects $\alpha=0.5$.
$\lambda$-Orth uses the paper objective, not an alternative code-objective
variant; 100 training epochs and predeclared epochs 1/10/25/50/100 give the
DEV-selected 100/25 epochs for Qwen/CLIP. MCT's complete mapping model
is trained with paired T2I positives and native MCL. Its selector maximizes
equal-dataset seven-point DEV raw R@1 AUC over two complete-model anchor
variants (T2I and bidirectional), temperatures $\{1,0.05\}$, and saved nonzero
epochs: $\{1,3,10\}$ at $T=1$ and $\{1,3,10,25,50\}$ at $T=0.05$.
Epoch 0 is diagnostic only. Ties prefer earlier epochs, then T2I; neither
selected winner is tied. Both winners use T2I and $T=0.05$: same-base Qwen selects epoch 25
with DEV AUC 70.4302\%, and CLIP epoch 50 with 60.5707\%. The epoch 25 checkpoint
is a prefix of the 50-epoch cosine schedule, not a fresh 25-epoch run. The
complete decision is frozen before selected TEST evaluation. The capacity
upgrade uses the same DEV AUC criterion over epochs 1, 3, 10, 25, and 50
for the paired T2I recipe. All main comparisons use complete MCT.
MKP reports a fixed 50-epoch CLIP representative and 10-epoch supplement.
FastFill reports fixed 80 epochs, with no epoch/coefficient search in the CLIP
run; XBT uses its fixed final 960-step CLIP adaptation. Their exact objectives
and recipes appear below. These development budgets differ; the comparisons
evaluate the method-specific recipes.

Earlier same-base MKP/XBT/FastFill records used an r16 teacher. The current
table uses separately completed r32 teacher runs for same-base and capacity
upgrades: MKP at 50 epochs, FastFill at 80 epochs, and full two-stage XBT.
CLIP RACT reuses the accepted fixed same-instance run. None of these
completions uses TEST for checkpoint selection. The RACT selection and
adapted objective are specified with the other baselines below.

\Needspace{6\baselineskip}
\subsection{Baseline implementations}
\label{sec:baseline-implementations}

\Needspace{4\baselineskip}
\subsubsection{Feature maps and scoring}
Let $q_o,d_o$ denote frozen old representations and $q_n,d_n$ the corresponding
Joint2880 teacher representations. Normalization is applied after mapping
unless stated otherwise. The following specifications describe our caption-retrieval implementations,
including the task adaptations and migration extensions used in this study.
In result tables, $^{*}$ marks RACT and FastFill, whose classification-based
objectives are adapted to paired retrieval as specified below.

\paragraph{RankMerge-basic.}
For an unrefreshed document use $\cos(q_o,d_o)$; for a refreshed document use
$\cos(q_n,d_n)$. Merge all candidates by that scalar score, with the common
document-row tie rule. This baseline merges raw scores directly. It requires both
query encoders while both subgalleries are nonempty. The endpoints are exactly
the old system and new Joint system. It is the basic versionwise merge control from the MCT study.

\Needspace{10\baselineskip}
\paragraph{SLERP with SVD alignment.}
Using row vectors here, accumulate $H=X_n^\top X_o$ from all paired TRAIN caption
and image draws (five text draws per image draw, no centering). Compute
$H=U\Sigma V^\top$ and $R=UV^\top$, without determinant correction. The query is
\[
 u_\alpha=\frac{\sin((1-\alpha)\theta)}{\sin\theta}q_o+
 \frac{\sin(\alpha\theta)}{\sin\theta}(q_nR),\qquad
 \theta=\arccos\langle q_o,q_nR\rangle.
\]
The cosine is clamped to $[-1+10^{-7},1-10^{-7}]$ before computing $\theta$; the output is normalized.
Old documents remain $d_o$ and refreshed documents use $\operatorname{norm}(d_nR)$.
$\alpha$ is fixed throughout migration. This compatible-document replay is an
explicit extension used here; pure $\alpha=1$ preserves the teacher endpoint.
SVD streams 4096-row FP32 feature blocks, text before image. It has no gradient
training or image re-encoding during fitting.

\paragraph{$\lambda$-Orth paper objective.}
$B$ is an affine $d\to d$ map with bias. The forward $F$ uses the original
DeepFCT-style two-branch MLP: each copy of old input traverses
$d\to 2d\to 2d$; their concatenation traverses $4d\to 2048\to 2048\to d $.
Hidden layers use batch normalization and ReLU; the final output is linear.
The two input branches both receive the old embedding, not extra side data.
With paired same-instance cross-model contrastive loss $C$ at logit scale 14,
the measured paper adaptation minimizes
\[
3\operatorname{MSE}(F(z_o),B(z_n))+
3\operatorname{MSE}(B(z_n),z_o)+
\tfrac12[C(F(z_o),B(z_n))+C(F(z_o),z_o)] + r\,\sigma(10(r-12)),
\]
where $r=\|WW^\top-I\|_F$ for $B$'s weight $W $.
The paired diagonal is a valid positive, not excluded as a within-view self-pair.
The additional $C(B(z_n),z_o)$ edge of an alternative code objective is not used
for the selected paper representative. Train on paired image and text records,
batch 4096, Adam $10^{-3}$, default betas/epsilon, no weight decay or schedule,
100 epochs with seeded per-epoch shuffle. The threshold 12 and forward
hidden width 2048 remain the same at 768D and 2048D.

At original-old-library entry, query $B(q_n)$ scores $d_o $. Conversion replaces
$d_o$ with $F(d_o)$, and subsequent re-encoding replaces that with $B(d_n)$;
query $B(q_n)$ stays fixed. Thus conversion end and re-encoding start share
exactly one gallery. The final endpoint still includes $B$, which can
alter the Joint ordering; it is not the unmapped teacher's endpoint.

\paragraph{Complete MCT.}
Both $\rho$ and $\psi$ use $d\to d\to d$ projections, with BatchNorm and ReLU
between affine layers. A shared $\rho$ processes concatenated new caption and
image features, preserving their 5:1 row ratio for batch normalization.
Set $a_j=\cos(\psi(\rho(q_n)),d_{o,j})/T$ and
$b_j=\cos(\rho(q_n),\rho(d_{n,j}))/T $. After mining the hard half of positives
and negatives separately for each version, write $P_a,P_b,N_a,N_b$ for their
sums of exponentiated logits. The two-term metric loss per query is
\[
-\log\frac{P_a}{P_a+N_a+N_b}
-\log\frac{P_b}{P_b+N_b+N_a}.
\]
For this caption task each query has one positive image; hardness keeps that
positive and the 64 hardest of 127 negative images in each version. Train with
128 image groups and all five captions per group, Adam $10^{-4}$ with cosine
annealing across 50 epochs, seeded epoch shuffle, no additional preservation
loss, and shared temperature $T=0.05$ selected on DEV. We use the native
old/new scores $a_jT$ and $b_jT$ at inference. Frozen teacher
representations are inputs, but the learned $\rho$ can change their ordering.
We implement this paired-T2I adaptation from the paper's formulation.

\paragraph{MKP adapter.}
At the shared retrieval dimension $d$, $h,g$ each use LayerNorm, affine $d\to d$,
GELU, dropout 0.1, and affine $d\to d $. Their used outputs are normalized.
The loss is mean coordinate-summed backward reconstruction plus equally
weighted new-feature point reconstruction, cosine-distance preservation, and
angle preservation, with focal exponent 1 on the three preservation terms.
Local relations use 100 exact full-training-population neighbors; global
relations use all nonself batch pairs; 32 global and 32 local angle tuples are
sampled per anchor. Batch 32, AdamW $10^{-3}$, weight decay $10^{-4}$, default
betas/epsilon, cosine over 50 epochs, fixed 50-epoch representative. The
inference query is $\operatorname{norm}(h(q_n))$; original document is unchanged.
Our compatible-document extension replaces it with $\operatorname{norm}(h(d_n))$.
This implementation uses normalized adapter outputs. The training-only $g$ is not an online encoder requirement.
For the two Qwen controls, $d=2048$ and the matching r32 Joint teacher is
frozen. Each completed 50-epoch fit performs 1,152,000 batch-32 updates over
737,280 cached query/document rows per epoch, with exact full-population
100-neighbor targets. The saved 10-epoch prefix is not the main representative.

\Needspace{12\baselineskip}
\paragraph{FastFill$^{*}$ retrieval adaptation (CLIP-to-Qwen).}
Let $f_i=F(d_{o,i})$ be the raw mapped document, $s_i$ its predicted scalar
log variance, $\widehat f_i=f_i/\|f_i\|$, and $\mathcal P_i$ the five frozen
new captions belonging to image $i $. For a global batch of $B=1024$ images
and $5B$ frozen new captions, the implemented losses are
\[
 \begin{aligned}
 \ell_{2,i}=\frac1d\|f_i-d_{n,i}\|_2^2,\qquad
 &\ell_{disc,i}=-\log
 \frac{\sum_{k\in\mathcal P_i}\exp(\widehat f_i^\top q_{n,k}/0.02)}
      {\sum_{k=1}^{5B}\exp(\widehat f_i^\top q_{n,k}/0.02)},\\
 &\mathcal L_{FF}=\frac1B\sum_i
 \left[\tfrac12 e^{-s_i}\left(\ell_{2,i}+\frac{\ell_{disc,i}}d\right)
       +\tfrac12 s_i\right],\quad d=768.
 \end{aligned}
\]
Thus discrimination sums positive probabilities, whereas the L2 term compares
\emph{raw} mapped vectors to unit teacher document targets. There is no category
classifier; the paired retrieval term replaces the original classifier-dependent
term. The released two-branch map has widths $d\to 2d\to 2d$ per branch,
concatenated into $4d\to 2048\to 2048\to d$, with BN/ReLU hidden layers and a
linear final output. Both branches receive the old document; an affine $d\to 1$
head on $f_i$ predicts unclipped $s_i $. No additional side information is used.

Train all map/head parameters in FP32 with Adam, learning rate $2\times10^{-5}$,
betas $(0.9,0.999)$, epsilon $10^{-8}$, weight decay $3.0517578125\times10^{-5}$,
and no gradient clipping. Eight workers each process 128 images, for 120 steps
per 122,880-image pass and 80 passes (9600 steps). Warmup is 600 steps,
followed by the cosine schedule. The fixed seeded row order repeats;
BN updates for the first 4800 steps and uses stored statistics thereafter,
while optimizer updates continue. The discrimination coefficient is fixed at
$1/768$, not selected on TEST. The final 9600-step checkpoint is fixed in advance.
At inference the frozen native Joint query scores normalized $F(d_o)$ during
conversion, then native $d_n$ during re-encoding. The primary refresh order is
the common hash order; descending predicted log variance is supplementary.
The raw query against unchanged $d_o$ is the original-gallery probe, not the
already-converted state.
The matched Qwen controls also use the fixed 80-epoch/9600-step endpoint
and their corresponding r32 Joint teacher. Both conversion and the subsequent
image re-encoding are measured; uncertainty-prioritized image refresh is
reported separately from the common-order primary comparison.

\paragraph{XBT retrieval adaptation (CLIP-to-Qwen).}
The shared projector $\phi$ has widths $d\to 4d\to 4d\to d$, LayerNorm/GELU
after each hidden affine layer, and unit-normalized input/output, with $d=768$.
Stage 1 fits all its parameters using the same 614,400 TRAIN captions encoded by
frozen CLIP and Joint query roles. Let
$\widehat q_{n,i}^{\epsilon}=\operatorname{norm}(q_{n,i}+\epsilon_i)$,
$\epsilon_i\sim\mathcal N(0,0.01^2I)$, and
$A_{ij}=e^{2.6592}\phi(\widehat q_{n,i}^{\epsilon})^\top q_{o,j}$.
With $\mathrm{CE}_{row}(A)$ denoting diagonal-label cross entropy,
\[
 \mathcal L_{X1}=\tfrac12\bigl[\mathrm{CE}_{row}(A)
                                  +\mathrm{CE}_{row}(A^\top)\bigr].
\]
Batch 8192 gives 75 steps for one fixed-order pass. AdamW uses learning rate
$10^{-4}$, weight decay 0.01, default betas/epsilon, and no scheduler. FP32 master
parameters use the BF16 autocast forwards. The fixed noise scale and
interpretation of 2.6592 as a \emph{log inverse temperature} are implementation
choices, not recovered author-code settings. No external 67M-caption corpus is used.

Stage 2 freezes the Joint teacher LoRAs and projector affine layers, adding
separate rank 32/alpha 64/dropout 0.05 query/document deltas and updating only those
deltas plus $\phi$'s LayerNorm parameters. For $G=128$ image groups, define
$S_{(gk),j}=e^{2.6592}\phi(Q(q_{gk}))^\top\phi(D(d_j))$ and use
\[
 \mathcal L_{X2}=\tfrac12\left[
 -\frac1{5G}\sum_{g,k}\log\frac{e^{S_{(gk),g}}}{\sum_j e^{S_{(gk),j}}}
 -\frac1{5G}\sum_{g,k}\log\frac{e^{S_{(gk),g}}}{\sum_{h,l}e^{S_{(hl),g}}}
 \right].
\]
The reverse term averages the five positive-caption log probabilities; it is
not the positive-probability sum used by FastFill. No old-model forward, L2,
MRL, or additional compatibility loss enters Stage 2. The run uses eight GPUs,
16 image groups per GPU, AdamW with betas $(0.9,0.95)$, epsilon $10^{-8}$,
weight decay 0.01, cosine/warmup ratio 0.001, and clipping 1. Delta LR is
$5\times10^{-4}$; projector LayerNorm LR is $10^{-4}$. The final 960-step
checkpoint is fixed, without DEV checkpoint search. Extra deltas add capacity:
combining each frozen rank 32 teacher with its rank 32 delta exports rank 64.
Original CLIP visual soft prompts are omitted for Qwen's token/grid layout.

For migration, projected adapted query scores unchanged CLIP Docs and
projected adapted Docs; there is no initial whole-gallery conversion. The
compatible endpoint retains $\phi$ and the learned deltas. Removing them
restores the independent teacher, which is a separately evaluated system,
not the compatible endpoint. The objectives and budgets above define the task adaptations evaluated here.
The matched Qwen controls use the full two-stage procedure at shared
dimension 2048: 75 text-projector updates followed by 960 extra-LoRA updates
over the corresponding r32 Joint teacher. The exported projector is applied
once by feature generation, with no second application in the scorer.

\Needspace{4\baselineskip}
\subsubsection{Same-instance \texorpdfstring{RACT$^{*}$}{RACT*}}
We adapt the compatibility term in Eq.~(6) of RACT~\cite{zhang2022ract} to
paired caption--image retrieval. We use Joint's grouped T2I MRL objective
as the main loss and apply the compatibility term within each modality.
It pairs the same input across model versions and uses both old and new
features as negatives.
For modality $m\in\{text,image\}$, let $a_i$ be a new anchor, $o_i$ its exact
same-input old feature, and $n_j$ a detached new-feature bank. Define
\[
 \ell_{m,i}=-\log\frac{e^{c(a_i,o_i)/0.05}}
 {e^{c(a_i,o_i)/0.05}+
  \sum_{j:\,I_j\ne I_i}\left(e^{c(a_i,\operatorname{sg}(o_j))/0.05}
       +e^{c(a_i,\operatorname{sg}(n_j))/0.05}\right)},
 \qquad
 \mathcal L=\mathcal L_{ret}+\lambda\bigl[\tfrac12\overline\ell_{text}
                              +\tfrac12\overline\ell_{image}\bigr].
\]
$c$ is cosine on the normalized shared-dimensional vector and $I$ is the canonical image
ID. A caption's paired old caption alone is the positive; its other same-image
captions are masked from negatives. Image anchors pair with their exact old
image. The modal means are global means; all old/new banks are detached, and
only local anchors reconnect through the original global-mean/DDP gradient
contract. Old query and document TRAIN features are therefore both required.
The Qwen main retrieval loss uses all four equally weighted MRL prefixes;
CLIP transfer uses the shared 768-dimensional embedding.

Both same-base candidates use seed 42, rank 32/alpha 64, batch 128 image groups with five
captions, eight GPUs, learning rate $5\times10^{-4}$, and the independent
full 2880-step Joint schedule. They do not initialize from a trained Joint.
DEV raw AUC is 70.276\% for $\lambda=0.1$ and 70.825\% for $\lambda=1$;
the predeclared rule chooses the latter (ties would use Doc100, then $\lambda=1$).
The decision is recorded before generating TEST features. There is no unselected
TEST evaluation, external score calibration, or stage 3. The main table and full
fraction records include the selected TEST curve; per-dataset endpoints
use the identical pools, relevance labels and exact scoring kernel.
The CLIP control fixes $\lambda=1$, compatibility temperature 0.05, main
temperature 0.02, rank 32/alpha 64/dropout 0.05, and 2880 steps at
LR $5\times10^{-4}$. Both roles are trained; its new paired endpoint is
69.85\%, rather than the independent Joint teacher's 73.15\%. It is an
exploratory task adaptation, without TEST-based parameter or checkpoint selection.

\Needspace{4\baselineskip}
\subsubsection{Migration extensions}
MKP learns $h$ from frozen new representations to the old space; its $g$ is
training-only. Our mapped-document migration evaluation applies $h$ to new document
features as well; this is an extension of its fixed-old-library evaluation,
not a guarantee in the original paper. SLERP's interpolated query similarly
uses a newly defined compatible-document replay here; the original appendix's
partial-backfill comparison does not imply this extension is guaranteed.
FastFill's classifier-dependent discrimination term is replaced by paired
retrieval supervision, and XBT lacks its original CLIP visual prompts.

MCT uses the complete $\rho+\psi+\mathrm{MCL}$ model. Its old query is $\psi(\rho(q))$,
new query is $\rho(q)$, and new document is $\rho(d)$. Its learned $\rho$ can change
the native teacher's ordering. In $\lambda$-Orth, the measured final system
retains $B$ on both query and document; soft orthogonality does not guarantee the
unchanged Joint endpoint. A raw Joint model remains an available
asset for these methods, but deploying it requires a corresponding document library
and an explicitly evaluated switch.

The evaluation maintains one active vector per document ID. A serving index
may also need storage for maintenance or rollback. MCT's disjoint version
subgalleries total $N$ active vectors as well. The logical vector count alone
therefore does not determine peak memory or serving latency.

\Needspace{6\baselineskip}
\subsection{Evaluation protocol}

\Needspace{4\baselineskip}
\subsubsection{Score calibration}
For fitting, DEV positive-image IDs are assigned by SHA-256(seed 20261107,ID)
modulo 2; one half is fit and one half held out diagnostically. This is a
held-out-query split, not an unseen-gallery split: the complete DEV gallery
remains available as negative candidates. Each query supplies one unit of
positive weight (shared across positives if needed), 50 highest-scoring
nonrelevant candidates, and 50 deterministic uniform nonrelevant samples.
Uniform negatives are sampled with replacement, excluding relevance targets
but allowing repetition and overlap with the hard-negative set; each drawn
occurrence retains its unit weight. Per-pool sample weights are divided by fit-query count, giving equal total
weight to COCO and Flickr. There is one map for each method/version state,
shared across both datasets and all fractions, and maps are locked before
TEST scoring. TEST labels, curves, and endpoints are not used to fit or select the maps.

Isotonic regression uses an increasing piecewise-linear interpolation, clipped
outside its fit-score range. A $10^{-9}$ raw-score refinement and raw-score
secondary ordering resolve plateaus before the common corpus-row tie rule.
The applied key is first the interpolated value plus $10^{-9}$ times raw
cosine, then raw cosine for any remaining exact floating tie, then original
corpus row. Floating ties are handled consistently across states. Raw replay, unchanged
pure endpoints, and exact conversion 100/re-encoding 0 seams are checked before
results are accepted. The 1:100 sampled prior is common across versions but
does not represent real-gallery relevance prevalence. No production
probability-calibration claim follows.

The fit subset contains 12,738 COCO and 2,570 Flickr queries. The held-out
subset contains 12,280 COCO and 2,500 Flickr queries and remains available for
encoder selection. In a diagnostic using earlier LoRA-only encoders, the
calibrated Qwen curve dipped below its entry score on this held-out subset.

\Needspace{4\baselineskip}
\subsubsection{Phase-specific AUCs}
Let $A_c,A_r$ be fraction-domain AUC for vector conversion and document
re-encoding. We report them separately: conversion writes $N$ vectors and
re-encoding writes another $N$. A direct method reports only $A_r$, beginning
with original vectors. A conversion method's $A_r$ instead begins with a
fully transformed gallery. Table~\ref{tab:projection-main} shows native
and calibrated $A_r$ for direct methods in panel A and conversion-first
methods in panel B. Panel B also reports $A_c$ and the fully converted
gallery's R@1, which starts the re-encoding phase.
An equal-time comparison would require conversion,
image encoding, index-update, and preparation timings at each fraction;
phase AUCs alone do not provide that comparison. Map training is preparation,
not conversion progress. The final Query switch changes no document vector and
is excluded from both phase AUCs.

\Needspace{4\baselineskip}
\subsubsection{Exact scoring}
Feature arrays are normalized in FP32. The evaluator computes dot products
in FP64 and resolves ties by original corpus row. Each method uses the
same corpus and query ID joins.

\Needspace{14\baselineskip}
\subsection{Computation and serving}

\Needspace{30\baselineskip}
\subsubsection{Method requirements}
These counts are encoder branches for a mixed gallery, not measured latency,
peak index storage, or complete GPU hours. Feature-map training, cached-feature
preparation, document updates and search operations remain separate work.
\begin{longtable}{p{0.15\textwidth}p{0.18\textwidth}p{0.29\textwidth}p{0.27\textwidth}}
\caption{Mixed-gallery requirements per method: query encoder branches, preparation and document work, and the mechanism that remains at the endpoint.}\label{tab:method-requirements}\\
\toprule Method & query encoder calls & Preparation / document work & final mechanism\\\midrule
\endfirsthead
\toprule Method & query encoder calls & Preparation / document work & final mechanism\\\midrule
\endhead
\addlinespace[2pt]
\multicolumn{4}{l}{\textit{Encoder compatibility}}\\*
RACT$^{*}$ & 1 new encoder & Old query/document TRAIN features; train both roles with compatibility; one document re-encoding pass & Native trained new pair\\
XBT & 1 new encoder plus projection & Joint teacher; text projection and additional Qwen LoRA & Retains compatible projection\\
BT$^{2*}$ & 1 new encoder, expanded by 32 & Frozen old and Joint teacher features; paired-retrieval student with cosine supervision; one document re-encoding pass; old vectors zero-padded during migration & Expanded student pair; the basis transformation stays inside the encoder\\
\addlinespace[2pt]
\multicolumn{4}{l}{\textit{Feature alignment}}\\*
SLERP & 1 at $\alpha=1$; 2 for $0<\alpha<1$ & Joint teacher, paired-feature fit, map each freshly encoded document & query interpolation and document rotation; pure SVD preserves native scores\\
MKP adapter & 1 new encoder, then $h$ & Joint teacher/features, relational targets; compatible document replay uses $h$ & Retains $h$; $g$ is training-only\\
\addlinespace[2pt]
\multicolumn{4}{l}{\textit{Versionwise retrieval}}\\*
RankMerge-basic & 2 until old subgallery empties & Independent new model; re-encode document; no learned merge map & Native new model; old branch retires\\
MCT & 1 new encoder; $\rho$ then $\psi$ for old branch & Joint teacher/features; train $\rho,\psi$; re-encode and apply $\rho$ & Retains $\rho$; $\psi$ retires after full refresh\\
\addlinespace[2pt]
\multicolumn{4}{l}{\textit{Two-pass conversion}}\\*
$\lambda$-Orth & 1 new encoder, then $B$ & Joint teacher/features; fit $B,F$; vector conversion then re-encoding & Retains $B$ on both query and document\\
FastFill$^{*}$ & 1 new encoder & Joint teacher/features; forward-map fit; vector conversion then document re-encoding & Native new pair after replacement\\
BiCT$^{*\dagger}$ & 1 new compatible encoder & Paired retrieval with a same-input cosine term; 900-step forward-map fit on old/new document features; vector conversion then our re-encoding extension & Compatible new pair; the conversion map retires once documents are re-encoded\\
Embedding-Converter$^{\dagger}$ & 1 frozen new encoder, no encoder training & Frozen source/target document features; 50,000-step converter fit; full vector conversion, then our re-encoding extension & Converted gallery served by the frozen new query; its own endpoint is full conversion\\
\addlinespace[2pt]
\multicolumn{4}{l}{\textit{Ours}}\\*
RELAY & 1 new query with optional projection & Query/Doc adaptation; optional query refinement; one document re-encoding pass & P100 keeps the query projection through backfill and removes it for refinement. The final pair uses fixed prefixes and adapted LoRAs.\\
\bottomrule\end{longtable}
An available raw teacher is an alternative model, not the measured compatible
endpoint. Removing a final map requires evaluating the matching document gallery
and switch. MCT's two disjoint version subgalleries total $N$ active vectors,
not two complete $N$-vector libraries. Encoder counts alone prove no complete
latency, memory, or preparation-cost advantage for RELAY.

\Needspace{4\baselineskip}
\subsubsection{Candidates for calibrated retrieval}
The merge in \cref{sec:deployment-work} uses the same
calibrated/raw/document-row key as exhaustive scoring. Version filtering
can use one filtered index or two disjoint subgalleries totaling $N$ active
vectors. The work includes up to two filtered searches and a merge from
one query embedding. Approximate search can omit candidates. Candidate
recall, latency, updates, and maintenance cost remain unmeasured.

\end{document}